\documentclass[runningheads]{llncs}
\usepackage{graphicx}
\usepackage{tikz}
\usepackage{comment} 
\usepackage{amsmath,amssymb} % define this before the line numbering.
\usepackage{color}

\usepackage{xcolor, soul}
\sethlcolor{pink}

\usepackage{xparse, mathtools, array}
\DeclarePairedDelimiterX{\rvect}[1]{[}{]}{\,\makervect{#1}\,}
\ExplSyntaxOn
\NewDocumentCommand{\makervect}{m}
 {
  \seq_set_split:Nnn \l_tmpa_seq { , } { #1 }
  \begin{matrix}
  \seq_use:Nn \l_tmpa_seq { & }
  \end{matrix}
 }
\ExplSyntaxOff
\newcommand{\Transp}{\mathsf{T}}

\newcommand{\parsection}[1]{\noindent\textbf{#1}~ }

\usepackage[title]{appendix}

\usepackage[pagebackref=true,breaklinks=true,colorlinks,bookmarks=false]{hyperref}
\usepackage{breakcites} % (needed for list of multiple citations to be line-breaked!)

\usepackage{booktabs}
\usepackage{multirow}

\usepackage{algorithm}
\usepackage[noend]{algpseudocode}
\makeatletter
\def\BState{\State\hskip-\ALG@thistlm}
\makeatother
\algnewcommand\Or{\textbf{or}}

\usepackage{amsmath,amsfonts,bm}

\def\eqref#1{equation~\ref{#1}}
\def\1{\bm{1}}

\newcommand{\test}{\mathcal{D_{\mathrm{test}}}}

\DeclareMathAlphabet{\mathsfit}{\encodingdefault}{\sfdefault}{m}{sl}
\SetMathAlphabet{\mathsfit}{bold}{\encodingdefault}{\sfdefault}{bx}{n}

\DeclareMathOperator*{\argmax}{arg\,max}

\usepackage[caption=false]{subfig}

\usepackage{pgfplots}
    \usepgfplotslibrary{colorbrewer}
    \pgfplotsset{
        cycle list/Dark2,
        cycle multiindex* list={
            mark list*\nextlist
            Dark2\nextlist
        },
    }
\pgfplotsset{compat=1.14}
\usepackage[width=122mm,left=12mm,paperwidth=146mm,height=193mm,top=12mm,paperheight=217mm]{geometry}

\begin{document}
% \linenumbers
\pagestyle{headings}
\mainmatter
\def\ECCVSubNumber{3472}  % Insert your submission number here

\title{Energy-Based Models for\\ Deep Probabilistic Regression} % Replace with your title

\begin{comment}
\titlerunning{ECCV-20 submission ID \ECCVSubNumber} 
\authorrunning{ECCV-20 submission ID \ECCVSubNumber} 
\author{Anonymous ECCV submission}
\institute{Paper ID \ECCVSubNumber}
\end{comment}
%******************

%\begin{comment}
\titlerunning{Energy-Based Models for Deep Probabilistic Regression}
\author{Fredrik K.~Gustafsson\inst{1} \and
Martin Danelljan\inst{2} \and\\
Goutam Bhat\inst{2} \and
Thomas B.~Sch\"on\inst{1}}
\authorrunning{F.K. Gustafsson et al.}
\institute{Department of Information Technology, Uppsala University, Sweden \and 
Computer Vision Lab, ETH Z\"urich, Switzerland}
%\end{comment}
%******************
\maketitle

\begin{abstract}
While deep learning-based classification is generally tackled using standardized approaches, a wide variety of techniques are employed for regression. In computer vision, one particularly popular such technique is that of confidence-based regression, which entails predicting a confidence value for each input-target pair $(x, y)$. While this approach has demonstrated impressive results, it requires important task-dependent design choices, and the predicted confidences lack a natural probabilistic meaning. We address these issues by proposing a general and conceptually simple regression method with a clear probabilistic interpretation. In our proposed approach, we create an energy-based model of the conditional target density $p(y | x)$, using a deep neural network to predict the un-normalized density from $(x, y)$. This model of $p(y | x)$ is trained by directly minimizing the associated negative log-likelihood, approximated using Monte Carlo sampling. We perform comprehensive experiments on four computer vision regression tasks. Our approach outperforms direct regression, as well as other probabilistic and confidence-based methods. Notably, our model achieves a $2.2\%$ AP improvement over Faster-RCNN for object detection on the COCO dataset, and sets a new state-of-the-art on visual tracking when applied for bounding box estimation. In contrast to confidence-based methods, our approach is also shown to be directly applicable to more general tasks such as age and head-pose estimation. Code is available at \url{https://github.com/fregu856/ebms_regression}.
\end{abstract}

\section{Introduction}
\label{section:introduction}

Supervised regression entails learning a model capable of predicting a continuous target value $y$ from an input $x$, given a set of paired training examples. It is a fundamental machine learning problem with many important applications within computer vision and other domains. Common regression tasks within computer vision include object detection~\cite{Ren2015FasterRT,jiang2018acquisition,law2018cornernet,zhou2019bottom}, head- and body-pose estimation~\cite{cao2017realtime,xiao2018simple,sun2019deep,yang2019fsa}, age estimation~\cite{rothe2016deep,pan2018mean,cao2019consistent}, visual tracking~\cite{MDNet,DaSiamRPN,SiamRPN++,danelljan2019atom} and medical image registration~\cite{niethammer2011geodesic,chou20132d}, just to mention a few. Today, such regression problems are commonly tackled using Deep Neural Networks (DNNs), due to their ability to learn powerful feature representations directly from data.

\begin{figure}[t]
\centering
   \includegraphics[width=0.790125\linewidth]{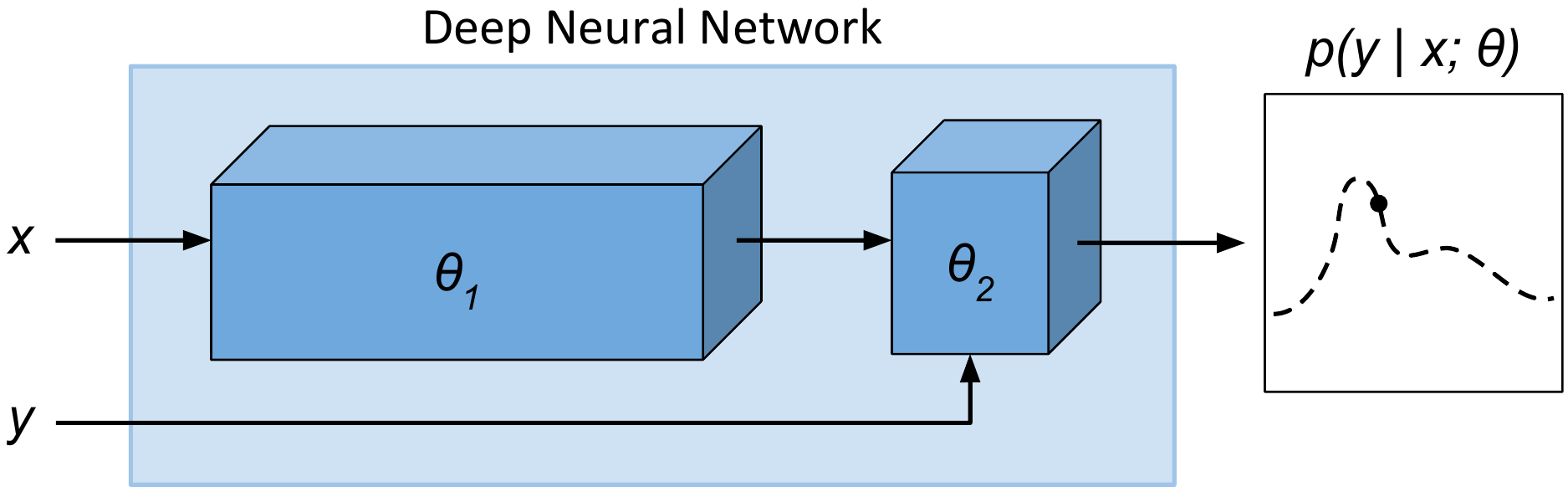}\vspace{-2mm}
   \begin{tabular}{@{}p{\columnwidth}@{}}
        \\\hline
   \end{tabular}\vspace{3mm}\\
   \newcommand{\imwid}{0.2595\columnwidth}%
   \includegraphics*[trim = 90 90 70 70, width=\imwid]{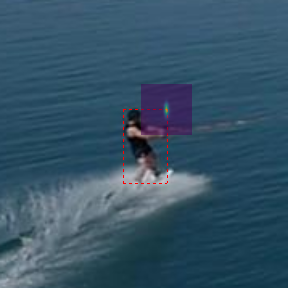}\hspace{2mm}%
   \includegraphics*[trim =  90 90 60 60, width=\imwid]{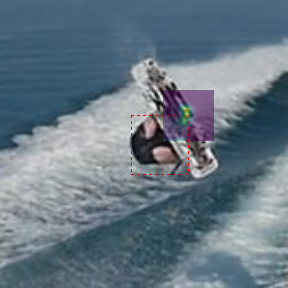}\hspace{2mm}%
   \includegraphics*[trim =  90 80 70 80, width=\imwid]{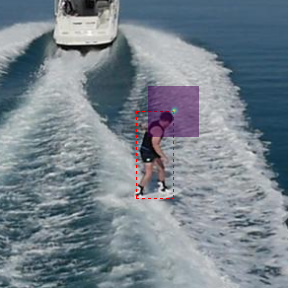}%\vspace{-2mm}%
   \caption{An overview of the proposed regression method (top). We train an energy-based model $p(y | x;\theta) \propto e^{f_\theta(x,y)}$ of the conditional target density $p(y | x)$, using a DNN $f_\theta$ to predict the un-normalized density directly from the input-target pair $(x, y)$. Our approach is capable of predicting highly flexible densities and produce highly accurate estimates. This is demonstrated for the problem of bounding box regression (bottom), visualizing the marginal density for the top right box corner as a heatmap.
   }%\vspace{-6mm}
\label{fig:intro_fig}
\end{figure}

While classification is generally addressed using standardized losses and output representations, a wide variety of different techniques are employed for regression. The most conventional strategy is to train a DNN to directly predict a target $y$ given an input $x$~\cite{lathuiliere2019comprehensive}. In such \emph{direct regression} approaches, the model parameters of the DNN are learned by minimizing a loss function, for example the $L^2$ or $L^1$ loss, penalizing discrepancy between the predicted and ground truth target values. From a probabilistic perspective, this approach corresponds to creating a simple parametric model of the conditional target density $p(y | x)$, and minimizing the associated negative log-likelihood. The $L^2$ loss, for example, corresponds to a fixed-variance Gaussian model. More recent work~\cite{kendall2017uncertainties,lakshminarayanan2017simple,chua2018deep,gast2018lightweight,varamesh2020mixture,prokudin2018deep} has also explored learning more expressive models of $p(y | x)$, by letting a DNN instead output the full set of parameters of a certain family of probability distributions. To allow for straightforward implementation and training, many of these \emph{probabilistic regression} approaches however restrict the parametric model to unimodal distributions such as Gaussian~\cite{lakshminarayanan2017simple,chua2018deep} or Laplace~\cite{kendall2017uncertainties,gast2018lightweight,ilg2018uncertainty}, still severely limiting the expressiveness of the learned conditional target density. While these methods benefit from a clear probabilistic interpretation, they thus fail to fully exploit the predictive power of the DNN. 

\begin{figure*}[ht]
\newcommand{\wid}{3.25cm}%
\newcommand{\imwid}{0.28\textwidth}%
\centering%
			\begin{tabular}{@{\hspace{-0.1cm}}c@{\hspace{1.8cm}}c@{\hspace{2.35cm}}c}
				Ground Truth & Gaussian & \textbf{Ours}
	\end{tabular}\vspace{0mm}
       \includegraphics[width=\imwid]{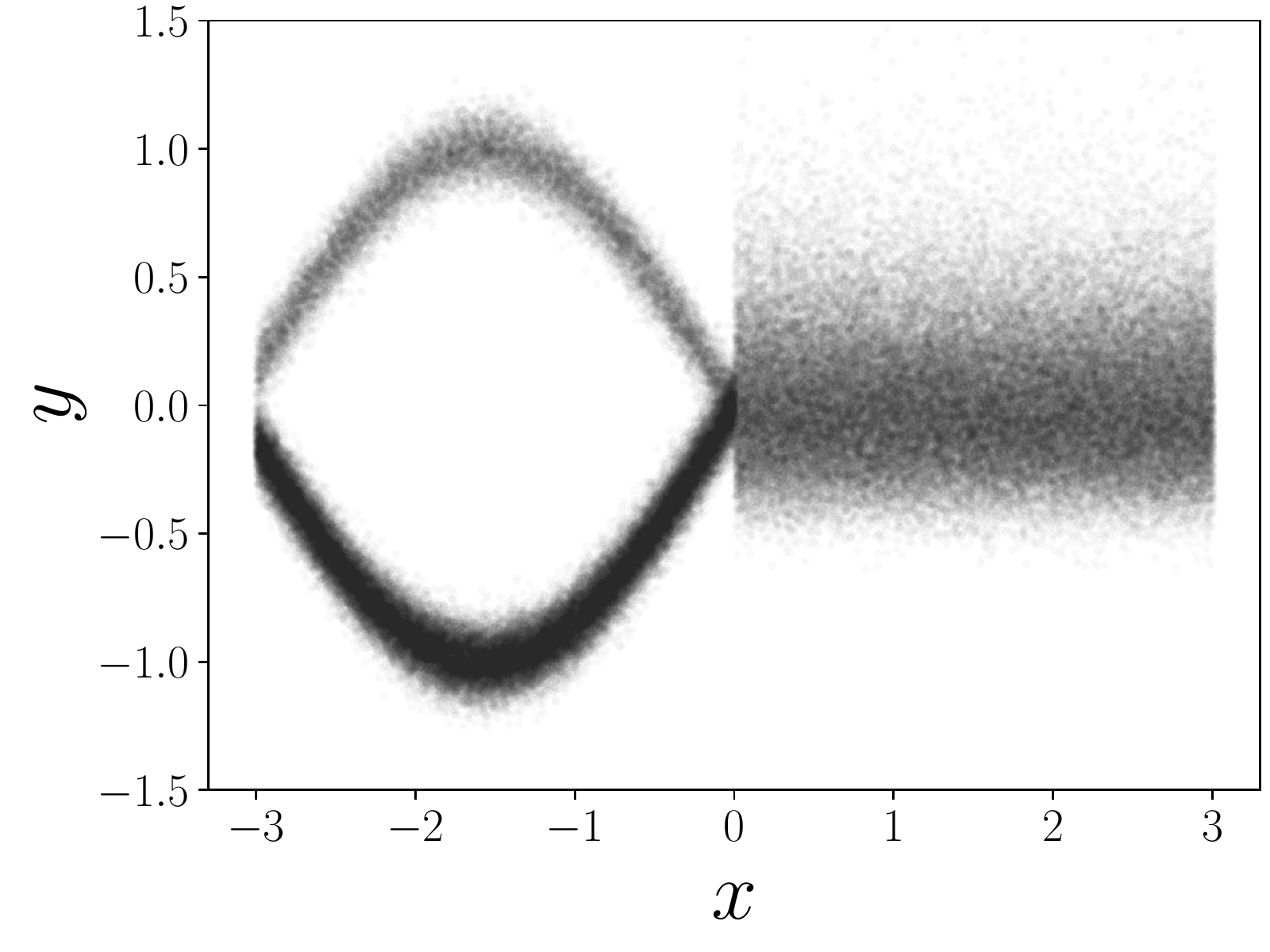}%
       \includegraphics[width=\imwid]{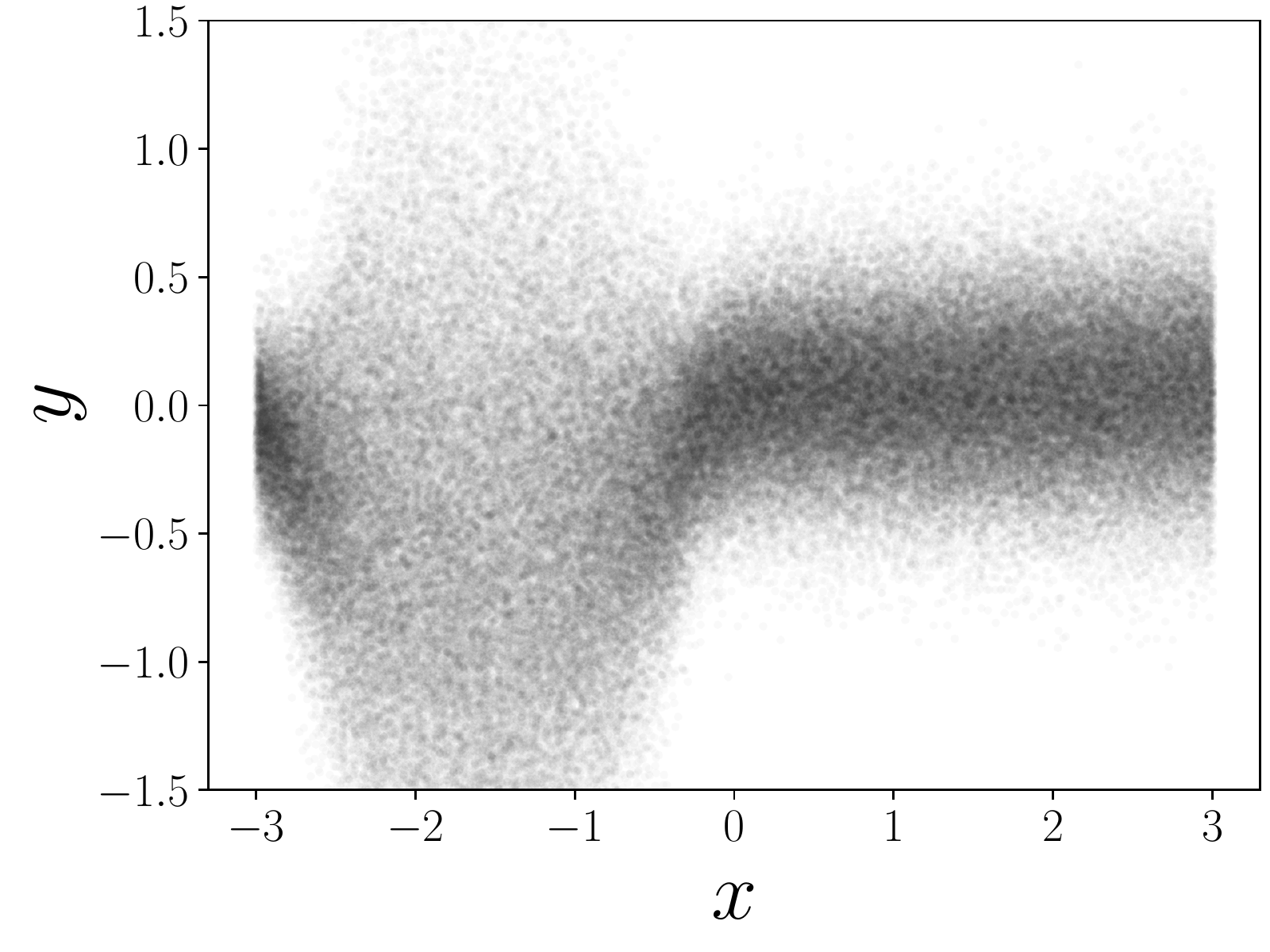}%
       \includegraphics[width=\imwid]{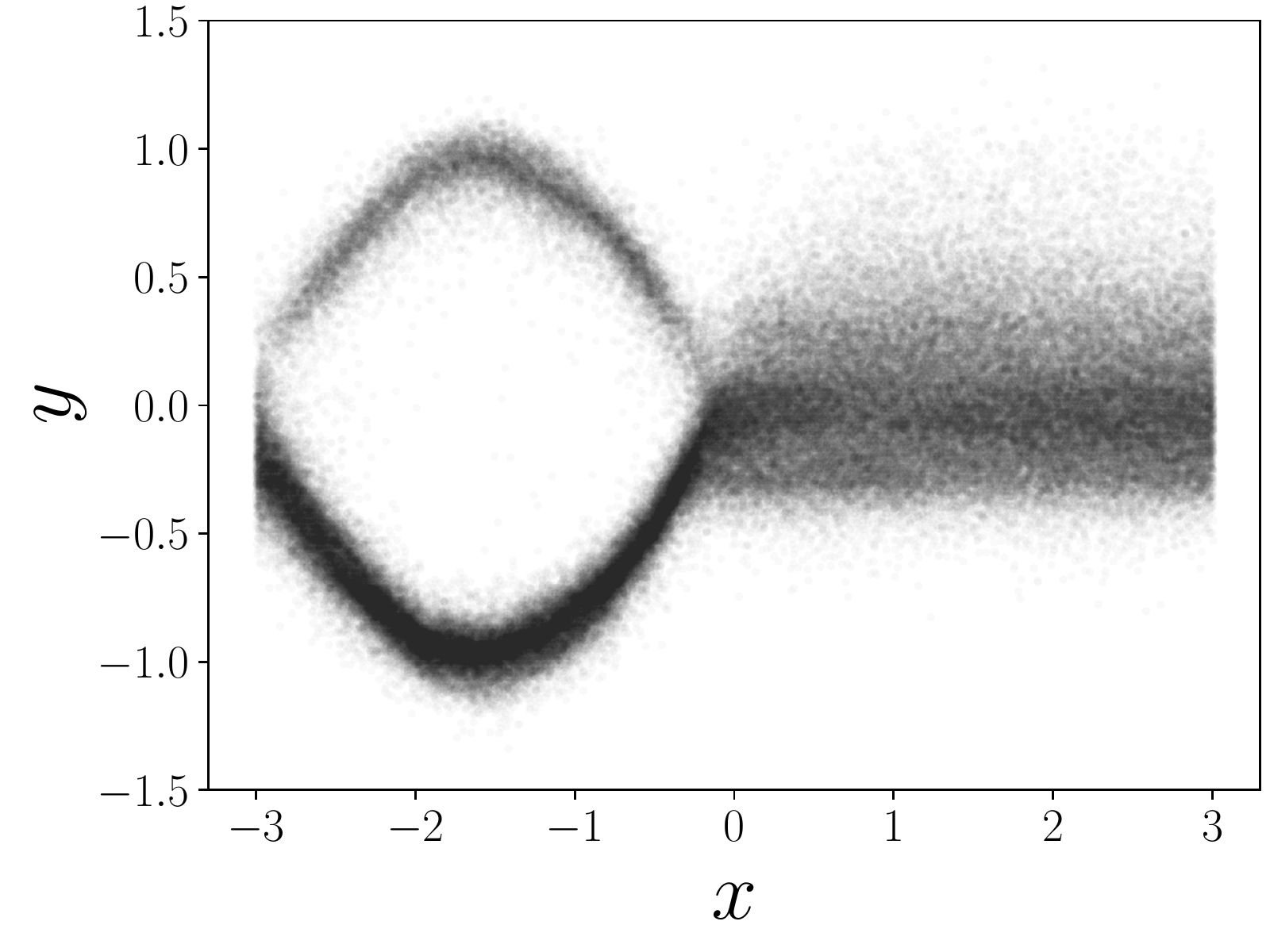}%
%\vspace{-4mm}
\caption{An illustrative 1D regression problem. The training data $\mathcal{D} = \{(x_i,y_i)\}_{i=1}^{2000}$ is generated by the ground truth conditional target density $p(y | x)$. Our energy-based model $p(y|x;\theta) \propto e^{f_\theta(x,y)}$ of $p(y | x)$ is trained by directly minimizing the associated negative log-likelihood, approximated using Monte Carlo importance sampling. In contrast to the Gaussian model $p(y | x; \theta)=\mathcal{N}(y;\mu_{\theta}(x),\sigma_{\theta}^2(x))$, our energy-based model can learn multimodal and complex conditional target densities directly from data.}%\vspace{-6mm}
\label{fig:overview}%
\end{figure*}

The quest for improved regression accuracy has also led to the development of more specialized methods, designed for a specific set of tasks. In computer vision, one particularly popular approach is that of \emph{confidence-based regression}. Here, a DNN instead predicts a scalar confidence value for input-target pairs $(x, y)$. The confidence can then be maximized w.r.t.\ $y$ to obtain a target prediction for a given input $x$. This approach is commonly employed for image-coordinate regression tasks within e.g.\ human pose estimation~\cite{cao2017realtime,xiao2018simple,sun2019deep} and object detection~\cite{law2018cornernet,zhou2019bottom}, where a 2D heatmap over image pixel coordinates $y$ is predicted. Recently, the approach was also applied to the problem of bounding box regression by Jiang et al.~\cite{jiang2018acquisition}. Their proposed method, IoU-Net, obtained state-of-the-art accuracy on object detection, and was later also successfully applied to the task of visual tracking~\cite{danelljan2019atom}. The training of such confidence-based regression methods does however entail generating additional pseudo ground truth labels, e.g.\ by employing a Gaussian kernel~\cite{convposeCVPR2016,xiao2018simple}, and selecting an appropriate loss function. This both requires numerous design choices to be made, and limits the general applicability of the methods. Moreover, confidence-based regression methods do not allow for a natural probabilistic interpretation in terms of the conditional target density $p(y | x)$. In this work, we therefore set out to develop a method combining the general applicability and the clear interpretation of probabilistic regression with the predictive power of the confidence-based approaches.

\parsection{Contributions} 
We propose a general and conceptually simple regression method with a clear probabilistic interpretation. Our method employs an energy-based model~\cite{lecun2006tutorial} to predict the un-normalized conditional target density $p(y | x)$ from the input-target pair $(x, y)$. It is trained by directly minimizing the associated negative log-likelihood, exploiting tailored Monte Carlo approximations. At test time, targets are predicted by maximizing the conditional target density $p(y | x)$ through gradient-based refinement. Our energy-based model is straightforward both to implement and train. Unlike commonly used probabilistic models, it can however still learn highly flexible target densities directly from data, as visualized in Figure~\ref{fig:overview}. Compared to confidence-based approaches, our method requires no pseudo labels, benefits from a clear probabilistic interpretation, and is directly applicable to a variety of computer vision applications. We evaluate the proposed method on four diverse computer vision regression tasks: object detection, visual tracking, age estimation and head-pose estimation. Our method is found to significantly outperform both direct regression baselines, and popular probabilistic and confidence-based alternatives, including the state-of-the-art IoU-Net~\cite{jiang2018acquisition}. Notably, our method achieves a $2.2\%$ AP improvement over FPN Faster-RCNN~\cite{lin2017feature} when applied for object detection on COCO~\cite{lin2014microsoft}, and sets a new state-of-the-art on standard benchmarks \cite{TrackingNet,UAV123} when applied for bounding box estimation in the recent ATOM~\cite{danelljan2019atom} visual tracker. Our method is also shown to be directly applicable to the more general tasks of age and head-pose estimation, consistently improving performance of a variety of baselines.
\section{Background \& Related Work}
\label{section:related_work}

In supervised regression, the task is to learn to predict a target value $y^\star \in \mathcal{Y}$ from a corresponding input $x^\star \in \mathcal{X}$, given a training set of i.i.d.\ input-target examples, $\mathcal{D} = \{(x_i, y_i)\}_{i=1}^{N}$, $(x_i, y_i) \sim p(x, y)$. As opposed to classification, the target space $\mathcal{Y}$ is a continuous set, e.g. $\mathcal{Y}=\mathbb{R}^K$. In computer vision, the input space $\mathcal{X}$ often corresponds to the space of images, whereas the output space $\mathcal{Y}$ depends on the task at hand. Common examples include $\mathcal{Y}=\mathbb{R}^2$ in image-coordinate regression~\cite{xiao2018simple,law2018cornernet}, $\mathcal{Y}=\mathbb{R}_+$ in age estimation~\cite{rothe2016deep,pan2018mean}, and $\mathcal{Y}=\mathbb{R}^4$ in object bounding box regression~\cite{Ren2015FasterRT,jiang2018acquisition}. A variety of techniques have previously been applied to supervised regression tasks. In order to motivate and provide intuition for our proposed method, we here describe a few popular approaches.

\parsection{Direct Regression}
Over the last decade, DNNs have been shown to excel at a wide variety of regression problems. Here, a DNN is viewed as a function $f_{\theta}: \mathcal{U} \rightarrow \mathcal{O}$, parameterized by a set of learnable weights $\theta \in \mathbb{R}^{P}$. The most conventional regression approach is to train a DNN to directly predict the targets, $y^\star = f_{\theta}(x^\star)$, called \emph{direct regression}. The model parameters $\theta$ are learned by minimizing a loss $\ell(f_\theta(x_i), y_i)$ that penalizes discrepancy between the prediction $f_\theta(x_i)$ and the ground truth target value $y_i$ on training examples $(x_i, y_i)$. Common choices include the $L^2$ loss, $\ell(\hat{y}, y) = \|\hat{y} - y\|^2_2$, the $L^1$ loss, $\ell(\hat{y},y) = \|\hat{y}-y\|_1$, and their close relatives~\cite{huber1964robust,lathuiliere2019comprehensive}. From a probabilistic perspective, the choice of loss corresponds to minimizing the negative log-likelihood $- \log p(y|x; \theta)$ for a specific model $p(y|x; \theta)$ of the conditional target density. For example, the $L^2$ loss is derived from a fixed-variance Gaussian model, $p(y|x; \theta) = \mathcal{N}(y; f_{\theta}(x),\sigma^2)$.

\parsection{Probabilistic Regression}
More recent work~\cite{kendall2017uncertainties,lakshminarayanan2017simple,chua2018deep,gast2018lightweight,ilg2018uncertainty,makansi2019overcoming,varamesh2020mixture} has explicitly taken advantage of this probabilistic perspective to achieve more flexible parametric models $p(y | x; \theta) = p(y; \phi_\theta(x))$, by letting the DNN output the parameters $\phi$ of a family of probability distributions $p(y; \phi)$. For example, a general 1D Gaussian model can be realized as $p(y | x; \theta) = \mathcal{N}\big(y; \mu_{\theta}(x),\,\sigma^2_{\theta}(x)\big)$, where the DNN outputs the mean and log-variance as $f_{\theta}(x)=\phi_{\theta}(x) = \rvect{\mu_{\theta}(x), \log\sigma^2_{\theta}(x)}^{\Transp} \in \mathbb{R}^2$. The model parameters $\theta$ are learned by minimizing the negative log-likelihood $- \sum_{i=1}^{N} \log p(y_i | x_i; \theta)$ over the training set $\mathcal{D}$. At test time, a target estimate $y^\star$ is obtained by first predicting the density parameter values $\phi_\theta(x^\star)$ and then, for instance, taking the expected value of $p(y; \phi_\theta(x))$. Previous work has applied simple Gaussian and Laplace models on computer vision tasks such as object detection~\cite{feng2019leveraging,he2019bounding} and optical flow estimation~\cite{gast2018lightweight,ilg2018uncertainty}, usually aiming to not only achieve accurate predictions, but also to provide an estimate of aleatoric uncertainty \cite{kendall2017uncertainties,gustafsson2020evaluating}. To allow for multimodal models $p(y; \phi_\theta(x))$, mixture density networks (MDNs) \cite{bishop1994mixture} have also been applied \cite{makansi2019overcoming,varamesh2020mixture}. The DNN then outputs weights for $K$ mixture components along with $K$ sets of parameters, e.g.\ $K$ sets of means and log-variances for a mixture of Gaussians. Previous work has also applied \emph{infinite} mixture models by utilizing the conditional VAE (cVAE) framework \cite{sohn2015learning,prokudin2018deep}. A latent variable model $p(y | x; \theta) = \int p(y; \phi_\theta(x, z)) p(z; \phi_\theta(x)) dz$ is then employed, where $p(y; \phi_\theta(x, z))$ and $p(z; \phi_\theta(x)$ typically are Gaussian distributions. Our proposed method also entails predicting a conditional target density $p(y | x; \theta)$ and minimizing the associated negative log-likelihood. However, our energy-based model $p(y | x; \theta)$ is not limited to the functional form of any specific probability density (e.g.\ Gaussian or Laplace), but is instead directly defined via a learned scalar function of $(x, y)$. In contrast to MDNs and cVAEs, our model $p(y | x; \theta)$ is not even limited to densities which are simple to generate samples from. This puts \emph{minimal restricting assumptions} on the true $p(y | x)$, allowing it to be efficiently learned directly from data.

\parsection{Confidence-Based Regression}
Another category of approaches reformulates the regression problem as $y^\star = \argmax_y f_\theta(x,y)$, where $f_\theta(x,y) \in \mathbb{R}$ is a scalar confidence value predicted by the DNN. The idea is thus to predict a quantity $f_\theta(x,y)$, depending on both input $x$ and target $y$, that can be maximized over $y$ to obtain the final prediction $y^\star$. This maximization-based formulation is inherent in  Structural SVMs \cite{SSVM}, but has also been adopted for DNNs. We term this family of approaches \emph{confidence-based regression}. Compared to direct regression, the predicted confidence $f_\theta(x,y)$ can encapsulate multiple hypotheses and other ambiguities. Confidence-based regression has been shown particularly suitable for image-coordinate regression tasks, such as hand keypoint localization \cite{simonCVPR2017} and body-part detection \cite{convposeCVPR2016,deepcutCVPR2016,xiao2018simple}. In these cases, a CNN is trained to output a 2D heatmap over the image pixel coordinates $y$, thus taking full advantage of the translational invariance of the problem. In computer vision, confidence prediction has also been successfully employed for tasks other than pure image-coordinate regression. Jiang et al.~\cite{jiang2018acquisition} proposed the IoU-Net for bounding box regression in object detection, where a bounding box $y \in \mathbb{R}^4$ and image $x$ are both input to the DNN to predict a confidence $f_\theta(x,y)$. It employs a pooling-based architecture that is differentiable w.r.t.\ the bounding box $y$, allowing efficient gradient-based maximization to obtain the final estimate $y^\star = \argmax_y f_\theta(x,y)$. IoU-Net was later also successfully applied to target object estimation in visual tracking \cite{danelljan2019atom}.

In general, confidence-based approaches are trained using a set of \emph{pseudo label} confidences $c(x_i, y_i, y)$ generated for each training example $(x_i, y_i)$, and by employing a loss $\ell\big(f_\theta(x_i, y), c(x_i, y_i, y)\big)$. One strategy \cite{deepcutCVPR2016,law2018cornernet} is to treat the confidence prediction as a binary classification problem, where $c(x_i, y_i, y)$ represents either the class, $c \in \{0,1\}$, or its probability, $c\in [0,1]$, and employ cross-entropy based losses $\ell$. The other approach is to treat the confidence prediction as a direct regression problem itself by applying standard regression losses, such as $L^2$ \cite{simonCVPR2017,danelljan2019atom,convposeCVPR2016} or the Huber loss \cite{jiang2018acquisition}. In these cases, the pseudo label confidences $c$ can be constructed using a similarity measure $S$ in the target space, $c(x_i, y_i, y) = S(y, y_i)$, for example defined as the Intersection over Union (IoU) between two bounding boxes \cite{jiang2018acquisition} or simply by a Gaussian kernel \cite{convposeCVPR2016,xiao2018simple,sun2019deep}.  

While these methods have demonstrated impressive results, confidence-based approaches thus require important design choices. In particular, the strategy for constructing the pseudo labels $c$ and the choice of loss $\ell$ are often crucial for performance and highly \emph{task-dependent}, limiting general applicability. Moreover, the predicted confidence $f_\theta(x,y)$ can be difficult to interpret, since it has no natural connection to the conditional target density $p(y | x)$. In contrast, our approach is directly trained to predict $p(y | x)$ itself, and importantly it does \textit{not} require generation of pseudo label confidences or choosing a specific loss.

\parsection{Regression-by-Classification}
A regression problem can also be treated as a classification problem by first discretizing the target space $\mathcal{Y}$ into a finite set of $C$ classes. Standard techniques from classification, such as softmax and the cross-entropy loss, can then be employed. This approach has previously been applied to both age estimation~\cite{rothe2016deep,pan2018mean,yang2018ssr} and head-pose estimation~\cite{ruiz2018fine,yang2019fsa}. The discretization of the target space $\mathcal{Y}$ however complicates exploiting its inherent neighborhood structure, an issue that has been addressed by exploring ordinal regression methods for 1D problems \cite{cao2019consistent,Diaz_2019_CVPR}. While our energy-based approach can be seen as a generalization of the softmax model for classification to the continuous target space $\mathcal{Y}$, it does not suffer from the aforementioned drawbacks of regression-by-classification. On the contrary, our model naturally allows the network to exploit the full structure of the continuous target space $\mathcal{Y}$.

\parsection{Energy-Based Models}
Our approach is of course also related to the theoretical framework of energy-based models, which often has been employed for machine learning problems in the past~\cite{mnih2005learning,hinton2006unsupervised,lecun2006tutorial}. It involves learning an energy function $\mathcal{E}_\theta(x) \in \mathbb{R}$ that assigns low energy to observed data $x_i$ and high energy to other values of $x$. Recently, energy-based models have been used primarily for unsupervised learning problems within computer vision~\cite{xie2016theory,gao2018learning,du2019implicit,lawson2019energy,nijkamp2019anatomy}, where DNNs are directly used to predict $\mathcal{E}_\theta(x)$. These models are commonly trained by minimizing the negative log-likelihood, stemming from the probabilistic model $p(x; \theta) = e^{-\mathcal{E}_\theta(x)}/\int e^{-\mathcal{E}_\theta(x)} dx$, for example by generating approximate image samples from $p(x; \theta)$ using Markov Chain Monte Carlo~\cite{gao2018learning,du2019implicit,nijkamp2019anatomy}. In contrast, we study the application of energy-based models for $p(y | x)$ in \emph{supervised} regression, a mostly overlooked research direction in recent years, and obtain state-of-the-art performance on four diverse computer vision regression tasks.
\section{Proposed Regression Method}
\label{section:DCTDs}

We propose a general and conceptually simple regression method with a clear probabilistic interpretation. Our method employs an energy-based model within a probabilistic regression formulation, defined in Section~\ref{DCTDs:formulation}. In Section~\ref{DCTDs:training}, we introduce our training strategy which is designed to be simple, yet highly effective and applicable to a wide variety of regression tasks within computer vision. Lastly, we describe our prediction strategy for high accuracy in Section~\ref{DCTDs:prediction}.

\subsection{Formulation}
\label{DCTDs:formulation}
We take the probabilistic view of regression by creating a model $p(y | x; \theta)$ of the conditional target density $p(y | x)$, in which $\theta$ is learned by minimizing the associated negative log-likelihood. Instead of defining $p(y | x; \theta)$ by letting a DNN predict the parameters of a certain family of probability distributions (e.g.\ Gaussian or Laplace), we construct a versatile energy-based model that can better leverage the predictive power of DNNs. To that end, we take inspiration from confidence-based regression approaches and let a DNN directly predict a scalar value for any input-target pair $(x, y)$. Unlike confidence-based methods however, this prediction has a clear probabilistic interpretation. Specifically, we view a DNN as a function $f_{\theta}: \mathcal{X} \times \mathcal{Y} \rightarrow \mathbb{R}$, parameterized by $\theta \in \mathbb{R}^{P}$, that maps an input-target pair $(x, y) \in \mathcal{X} \times \mathcal{Y}$ to a scalar value $f_{\theta}(x, y) \in \mathbb{R}$. Our model $p(y | x; \theta)$ of the conditional target density $p(y | x)$ is then defined according to,
\begin{equation}
    p(y | x; \theta) = \frac{e^{f_{\theta}(x, y)}}{Z(x, \theta)}, \qquad Z(x, \theta) = \int e^{f_{\theta}(x, \tilde{y})} d\tilde{y} \,,
\label{eq:DCTD_def}
\end{equation}
where $Z(x, \theta)$ is the input-dependent normalizing partition function. We train this energy-based model (\ref{eq:DCTD_def}) by directly minimizing the negative log-likelihood $-\log p(\{y_i\}_i | \{x_i\}_i; \theta) = \sum_{i=1}^{N} - \log p(y_i | x_i; \theta)$, where each term is given by,
\begin{equation}
    -\log p(y_i | x_i; \theta) = \log \bigg(\int e^{f_{\theta}(x_i, y)} dy\bigg) - f_{\theta}(x_i, y_i).
\label{eq:nll}
\end{equation}
This direct and straightforward training approach thus requires the evaluation of the generally intractable $Z(x, \theta) = \int e^{f_{\theta}(x, y)} dy$. Many fundamental computer vision tasks, such as object detection, keypoint estimation and pose estimation, however rely on regression problems with a low-dimensional target space $\mathcal{Y}$. In such cases, effective finite approximations of $Z(x, \theta)$ can be applied. In some tasks, such as image-coordinate regression, this is naturally performed by a grid approximation, utilizing the dense prediction obtained by fully-convolutional networks. In this work, we however investigate a more \emph{generally applicable} technique, namely Monte Carlo approximations with importance sampling. This procedure, when employed for training the network, is detailed in Section~\ref{DCTDs:training}.

At test time, given an input $x^\star$, our model in (\ref{eq:DCTD_def}) allows evaluating the conditional target density $p(y|x^\star; \theta)$ for any target $y$ by first approximating $Z(x^\star, \theta)$, and then predicting the scalar $f_\theta(x^\star, y)$ using the DNN. This enables the computation of, e.g., the mean and variance of the target value $y$. In this work, we take inspiration from confidence-based regression and focus on finding the most likely prediction, $y^\star = \argmax_y p(y | x^\star; \theta) = \argmax_y f_\theta(x^\star, y)$, which does not require the evaluation of $Z(x^\star, \theta)$ during inference. Thanks to the auto-differentiation capabilities of modern deep learning frameworks, we can apply gradient-based techniques to find the final prediction by simply maximizing the network output $f_\theta(x^\star, y)$ w.r.t.\ $y$. We elaborate on this procedure for prediction in Section~\ref{DCTDs:prediction}.

%%%%%%%%%%%%%%%%%%%%%%%%%%%%%%%%%%%%%%%%%%%%%%%%%%%%%%%%%%%%%%%%%%%%%%%%%%%%%%%%%%%%
% TRAINING:
%%%%%%%%%%%%%%%%%%%%%%%%%%%%%%%%%%%%%%%%%%%%%%%%%%%%%%%%%%%%%%%%%%%%%%%%%%%%%%%%%%%%
\subsection{Training}
\label{DCTDs:training}

Our model $p(y | x; \theta) = e^{f_{\theta}(x, y)}/Z(x, \theta)$ of the conditional target density is trained by directly minimizing the negative log-likelihood $\sum_{i=1}^{N} -\log p(y_i | x_i; \theta)$. To evaluate the integral in (\ref{eq:nll}), we employ Monte Carlo importance sampling. Each term $-\log p(y_i | x_i; \theta)$ is therefore approximated by sampling values $\{y^{(k)}\}_{k=1}^{M}$ from a proposal distribution $q(y|y_i)$ that depends on the ground truth target value $y_i$, 
\begin{equation}
    -\log p(y_i| x_i;\theta) \approx \log \bigg(\frac{1}{M} \sum_{k=1}^{M} \frac{e^{f_{\theta}(x_i, y^{(k)})}}{q(y^{(k)}|y_i)} \bigg) - f_{\theta}(x_i, y_i).
\label{eq:negative_log_likelihood_term}
\end{equation}
The final loss $J(\theta)$ used to train the DNN $f_{\theta}$ is then obtained by averaging over all training examples $\{(x_i,y_i)\}_{i=1}^{n}$ in the current mini-batch,
\begin{equation}
\label{eq:djsp_loss_function}
    J(\theta) = \frac{1}{n} \sum_{i = 1}^{n} \log \bigg(\frac{1}{M} \sum_{m=1}^{M} \frac{e^{f_{\theta}(x_i, y^{(i, m)})}}{q(y^{(i, m)}|y_i)} \bigg) - f_{\theta}(x_i, y_i),
\end{equation}
where $\{y^{(i, m)}\}_{m=1}^{M}$ are $M$ samples drawn from $q(y|y_i)$. Qualitatively, minimizing $J(\theta)$ encourages the DNN to output large values $f_{\theta}(x_i, y_i)$ for the ground truth target $y_i$, while minimizing the predicted value $f_{\theta}(x_i, y)$ at all other targets $y$. In ambiguous or uncertain cases, the DNN can output small values everywhere or large values at multiple hypotheses, but at the cost of a higher loss.

As can be seen in (\ref{eq:djsp_loss_function}), the DNN $f_{\theta}$ is applied both to the input-target pair $(x_i, y_i)$, and all input-sample pairs $\{(x_i, y^{(i, m)})\}_{m=1}^{M}$ during training. While this can seem inefficient, most applications in computer vision employ network architectures that first extract a deep feature representation for the input $x_i$. The DNN $f_{\theta}$ can thus be designed to combine this input feature with the target $y$ at a late stage, as visualized in Figure~\ref{fig:intro_fig}. The input feature extraction process, which becomes the main computational bottleneck, therefore needs to be performed only once for each $x_i$. In practice, we found our training strategy to not add any significant overhead compared to the direct regression baselines, and the computational cost to be \emph{identical} to that of the confidence-based methods.

Compared to confidence-based regression, a significant advantage of our approach is however that there is no need for generating task-dependent pseudo label confidences or choosing between different losses. The \emph{only} design choice of our training method is the proposal distribution $q(y|y_i)$. Note however that since the loss $J(\theta)$ in (\ref{eq:djsp_loss_function}) explicitly adapts to $q(y|y_i)$, this choice has no effect on the overall behaviour of the loss, only on the quality of the sampled approximation. We found a mixture of a few equally weighted Gaussian components, all centered at the target label $y_i$, to consistently perform well in our experiments across all four diverse computer vision applications. Specifically, $q(y|y_i)$ is set to,
\begin{equation}
\label{eq:proposal_distribution}
    q(y|y_i) = \frac{1}{L} \sum_{l=1}^{L} \mathcal{N}(y; y_i, \sigma_{l}^{2}I),
\end{equation}
where the standard deviations $\{\sigma_{l}\}_{l=1}^{L}$ are hyperparameters selected based on a validation set for each experiment. We only considered the simple Gaussian proposal in (\ref{eq:proposal_distribution}), as this was found sufficient to obtain state-of-the-art experimental results. Full ablation studies for the number of components $L$ and $\{\sigma_{l}\}_{l=1}^{L}$ are provided in the supplementary material. Figure~\ref{fig:overview} illustrates that our model $p(y | x; \theta)$ can learn complex conditional target densities, containing both multi-modalities and asymmetry, directly from data using the described training procedure. In this illustrative example, we use (\ref{eq:proposal_distribution}) with $L=2$ and $\sigma_1 = 0.1$, $\sigma_2 = 0.8$.

%%%%%%%%%%%%%%%%%%%%%%%%%%%%%%%%%%%%%%%%%%%%%%%%%%%%%%%%%%%%%%%%%%%%%%%%%%%%%%%%%%%%
% PREDICTION:
%%%%%%%%%%%%%%%%%%%%%%%%%%%%%%%%%%%%%%%%%%%%%%%%%%%%%%%%%%%%%%%%%%%%%%%%%%%%%%%%%%%%
\subsection{Prediction}
\label{DCTDs:prediction}

Given an input $x^\star$ at test time, the trained DNN $f_\theta$ can be used to evaluate the full conditional target density $p(y | x^\star; \theta) = e^{f_{\theta}(x^\star, y)}/Z(x^\star, \theta)$, by employing the aforementioned techniques to approximate the constant $Z(x^\star, \theta)$. In many applications, the most likely prediction $y^\star = \argmax_y p(y | x^\star; \theta)$ is however the single desired output. For our energy-based model, this is obtained by directly maximizing the DNN output, $y^\star = \argmax_y f_\theta(x^\star, y)$, thus not requiring $Z(x^\star, \theta)$ to be evaluated. By taking inspiration from IoU-Net~\cite{jiang2018acquisition} and designing the DNN $f_\theta$ to be differentiable w.r.t.\ the target $y$, the gradient $\nabla_{y} f_{\theta}(x^\star, y)$ can be efficiently evaluated using the auto-differentiation tools implemented in modern deep learning frameworks. An estimate of $y^\star = \argmax_y f_\theta(x^\star, y)$ can therefore be obtained by performing gradient ascent to find a local maximum of $f_\theta(x^\star, y)$. 

The gradient ascent refinement is performed either on a single initial estimate $\hat{y}$, or on a set of random initializations $\{\hat{y}_k\}_{k=1}^K$ to obtain a final accurate prediction $y^\star$. Starting at $y = \hat{y}_k$, we thus run $T$ gradient ascent iterations, $y \gets y + \lambda \nabla_{y} f_{\theta}(x^\star, y)$, with step-length $\lambda$. In our experiments, we fix $T$ (typically, $T\!=\!10$) and select $\lambda$ using grid search on a validation set. As noted in Section~\ref{DCTDs:training}, this prediction procedure can be made highly efficient by extracting the feature representation for $x^\star$ only once. Back-propagation is then performed only through a few final layers of the DNN to evaluate the gradient $\nabla_{y} f_{\theta}(x^\star, y)$. The gradient computation for a set of candidates $\{\hat{y}_k\}_{k=1}^K$ can also be parallelized on the GPU by simple batching, requiring no significant overhead. Overall, the inference speed is somewhat decreased compared to direct regression baselines, but is \emph{identical} to confidence-based methods such as IoU-Net~\cite{jiang2018acquisition}. An algorithm detailing this prediction procedure is found in the supplementary material.
\section{Experiments}
\label{section: experiments}

We perform comprehensive experiments on four different computer vision regression tasks: object detection, visual tracking, age estimation and head-pose estimation. Our proposed approach is compared both to baseline regression methods and to state-of-the-art models. Notably, our method significantly outperforms the confidence-based IoU-Net~\cite{jiang2018acquisition} method for bounding box regression in direct comparisons, both when applied for object detection on the COCO dataset~\cite{lin2014microsoft} and for target object estimation in the recent ATOM~\cite{danelljan2019atom} visual tracker. On age and head-pose estimation, our approach is shown to consistently improve performance of a variety of baselines. All experiments are implemented in PyTorch~\cite{paszke2019pytorch}. For all tasks, further details are also provided in the supplementary material.

\begin{table*}[t]
\centering
\caption{Impact of $L$ and $\{\sigma_{l}\}_{l=1}^{L}$ in the proposal distribution $q(y|y_i)$~(\ref{eq:proposal_distribution}), for the object detection task on the \emph{2017 val} split of the COCO~\cite{lin2014microsoft} dataset. For $L=2$, $\sigma_{1} = \sigma_{2}/4$. For $L=3$, $\sigma_{1} = \sigma_{3}/4$ and $\sigma_{2} = \sigma_{3}/2$. $L=3$ with $\sigma_{L}=0.15$ is selected.}
\resizebox{0.7825\textwidth}{!}{%
\begin{tabular}{l@{\hspace{1cm}}c@{~~}c@{~~}c@{~~}@{~~}c@{~~}c@{~~}c@{~~}@{~~}c@{~~}c@{~~}c}
\toprule
Number of components $L$ & &1 & & &2 & & &3 &\\
Base proposal st.\ dev.\ $\sigma_{L}$ &0.02 &0.04 &0.08 &0.1 &0.15 &0.2 &0.1 &0.15 &0.2\\
\midrule
AP (\%) &38.1 &38.5 &37.5 &39.0 &\textbf{39.1} &39.0 &39.0 &\textbf{39.1} &38.8\\
\bottomrule
\end{tabular}}%\vspace{-3.0mm}
\label{table:detection_ablation}%
\end{table*}

\begin{table*}[t]
\centering
	\caption{Results for the object detection task on the \emph{2017 test-dev} split of the COCO~\cite{lin2014microsoft} dataset. Our proposed method significantly outperforms the baseline FPN Faster-RCNN~\cite{lin2017feature} and the state-of-the-art confidence-based IoU-Net~\cite{jiang2018acquisition}.}%\vspace{-3mm}
	\resizebox{1.0\textwidth}{!}{%
		\begin{tabular}{l@{\hspace{0.5cm}}cccccccccc}
\toprule
Formulation &Direct &Gaussian &Gaussian &Gaussian &Gaussian &Gaussian &Laplace &Confidence &Confidence &\textbf{}\\ 
Approach &Faster-RCNN & &Mixt. 2 &Mixt. 4 &Mixt. 8 &cVAE & &IoU-Net &IoU-Net$^*$ &\textbf{Ours}\\
\midrule
AP (\%)&37.2 &36.7 &37.1 &37.0 &36.8 &37.2 &37.1 &38.3 &38.2 &\textbf{39.4}\\
AP$_\text{50} (\%)$&\textbf{59.2} &58.7 &59.1 &59.1 &59.1 &\textbf{59.2} &59.1 &58.3 &58.4 &58.6\\
AP$_\text{75} (\%)$&40.3 &39.6 &40.0 &39.9 &39.7 &40.0 &40.2 &41.4 &41.4 &\textbf{42.1}\\
FPS &\textbf{12.2} &\textbf{12.2} &\textbf{12.2} &12.1 &12.1 &9.6 &\textbf{12.2} &5.3 &5.3 &5.3\\
\bottomrule
\end{tabular}
	}%\vspace{-3.0mm}
	\label{tab:detection}
\end{table*}

%%%%%%%%%%%%%%%%%%%%%%%%%%%%%%%%%%%%%%%%%%%%%%%%%%%%%%%%%%%%%%%%%%%%%%%%%%%%%%%
\subsection{Object Detection}
\label{experiments:object_detection}
We first perform experiments on object detection, the task of classifying and estimating a bounding box for each object in a given image. Specifically, we compare our regression method to other techniques for the task of bounding box regression, by integrating them  into an existing object detection pipeline. To this end, we use the Faster-RCNN \cite{Ren2015FasterRT} framework, which serves as a popular baseline in the object detection field due to its strong state-of-the-art performance. It employs one network head for classification and one head for regressing the bounding box using the direct regression approach. We also include various probabilistic regression baselines and compare with simple Gaussian and Laplace models, by modifying the Faster-RCNN regression head to predict both the mean and log-variance of the distribution, and adopting the associated negative log-likelihood loss. Similarly, we compare with mixtures of $K = \{2, 4, 8\}$ Gaussians by duplicating the modified regression head $K$ times and adding a network head for predicting $K$ component weights. Moreover, we compare with an infinite mixture of Gaussians by training a cVAE. Finally, we also compare our approach to the state-of-the-art confidence-based IoU-Net \cite{jiang2018acquisition}. It extends Faster-RCNN with an additional branch that predicts the IoU overlap between a target bounding box $y$ and the ground truth. The IoU prediction branch uses differentiable region pooling \cite{jiang2018acquisition}, allowing the initial bounding box predicted by the Faster-RCNN to be refined using gradient-based maximization of the predicted IoU confidence. 

For our approach, we employ an \emph{identical architecture} as used in IoU-Net for a fair comparison. Instead of training the network to output the IoU, we predict the exponent $f_\theta(x,y)$ in (\ref{eq:DCTD_def}), trained by minimizing the negative log-likelihood in (\ref{eq:djsp_loss_function}). We parametrize the bounding box as $y = (c_x/w_0, c_y/h_0, \log w, \log h) \in \mathbb{R}^4$, where $(c_x, c_y)$ and $(w,h)$ denote the center coordinate and size, respectively. The reference size $(w_0, h_0)$ is set to that of the ground truth during training and the initial box during prediction. Based on the ablation study found in Table~\ref{table:detection_ablation}, we employ $L=3$ isotropic Gaussians with standard deviation $\sigma_l = 0.0375 \cdot 2^{l-1}$ for the proposal distribution (\ref{eq:proposal_distribution}). In addition to the standard IoU-Net, we compare with a version (denoted IoU-Net$^*$) employing the same proposal distribution and inference settings as in our approach. For both our method and IoU-Net$^*$, we set the refinement step-length $\lambda$ using grid search on a separate validation set.

Our experiments are performed on the large-scale COCO benchmark~\cite{lin2014microsoft}. We use the \emph{2017 train} split ($\approx$ 118\thinspace000 images) for training and the \emph{2017 val} split ($\approx$ 5\thinspace000 images) for setting our hyperparameters. The results are reported on the \emph{2017 test-dev} split ($\approx$  20\thinspace000 images), in terms of the standard COCO metrics AP,  AP$_\text{50}$ and AP$_\text{75}$. We also report the inference speed in terms of frames-per-second (FPS) on a single NVIDIA TITAN Xp GPU. We initialize all networks in our comparison with the pre-trained Faster-RCNN weights, using the ResNet50-FPN \cite{lin2017feature} backbone, and re-train \emph{only} the newly added layers for a fair comparison. The results are shown in Table~\ref{tab:detection}. Our proposed method obtains the best results, significantly outperforming Faster-RCNN and IoU-Net by $2.2\%$ and $1.1\%$ in AP, respectively. The Gaussian model is outperformed by the mixture of $2$ Gaussians, but note that adding more components does \emph{not} further improve the performance. In comparison, the cVAE achieves somewhat improved performance, but is still clearly outperformed by our method. Compared to the probabilistic baselines, we believe that our energy-based model offers a more direct and effective representation of the underlying density via the scalar DNN output $f_{\theta}(x, y)$. The inference speed of our method is somewhat lower than that of Faster-RCNN, but identical to IoU-Net. How the number of iterations $T$ in the gradient-based refinement affects inference speed and performance is analyzed in Figure~\ref{fig:detection_impact_of_grad_iters}, showing that our choice $T=10$ provides a reasonable trade-off.

\begin{table*}[t]
\centering
	\caption{Results for the visual tracking task on the two common datasets TrackingNet~\cite{TrackingNet} and UAV123~\cite{UAV123}. The symbol $^\dagger$ indicates an approximate value ($\pm 1)$, taken from the plot in the corresponding paper. Our proposed method significantly outperforms the baseline ATOM and other recent state-of-the-art trackers.}%\vspace{0mm}
	\resizebox{1.0\textwidth}{!}{%
		\begin{tabular}{ll@{~~~}c@{~~~}c@{~~~}c@{~~~}c@{~~~}c@{~~~}c@{~~~}c@{~~~}c@{~~~}c}
	\toprule
	\multirow{2}{14mm}{Dataset}&\multirow{2}{14mm}{Metric}&ECO&SiamFC&MDNet&UPDT&DaSiamRPN&SiamRPN++&ATOM&ATOM$^*$&\textbf{Ours}\\
	&&\cite{DanelljanCVPR2017}&\cite{SiameseFC}&\cite{MDNet}&\cite{bhat2018unveiling}&\cite{DaSiamRPN}&\cite{SiamRPN++}& \cite{danelljan2019atom}&&\\
	\midrule
	\multirow{3}{19mm}{TrackingNet}
	&Precision (\%)&49.2&53.3&56.5&55.7&59.1&69.4&64.8&66.6&\textbf{69.7}\\
	&Norm.\ Prec.\ (\%)&61.8&66.6&70.5&70.2&73.3&80.0&77.1&78.4&\textbf{80.1}\\
	&Success (\%)&55.4&57.1&60.6&61.1&63.8&73.3&70.3&72.0&\textbf{74.5}\\
	\midrule
	\multirow{3}{19mm}{UAV123}
	&OP$_{0.50}$ (\%)&64.0&-&-&66.8&73.6&75$^\dagger$&78.9&79.0&\textbf{80.8}\\
	&OP$_{0.75}$ (\%)&32.8&-&-&32.9&41.1&56$^\dagger$&55.7&56.5&\textbf{60.2}\\
	&AUC (\%)&53.7&-&52.8&55.0&58.4&61.3&65.0&64.9&\textbf{67.2}\\
	\bottomrule
\end{tabular}

	}%\vspace{-4.0mm}
	\label{tab:tracking}%
\end{table*}

%%%%%%%%%%%%%%%%%%%%%%%%%%%%%%%%%%%%%%%%%%%%%%%%%%%%%%%%%%%%%%%%%%%%%%%%%%%%%%%
\subsection{Visual Tracking}
\label{experiments:visual_tracking}

Next, we evaluate our approach on the problem of generic visual object tracking. The task is to estimate the bounding box of a target object in every frame of a video. The target object is defined by a given box in the first video frame. We employ the recently introduced ATOM \cite{danelljan2019atom} tracker as our baseline. Given the first-frame annotation, ATOM trains a classifier to first roughly localize the target in a new frame. The target bounding box is then determined using an IoU-Net-based module, which is also conditioned on the first-frame target appearance using a modulation-based architecture. We train our network to predict the conditional target density through $f_\theta(x,y)$ in (\ref{eq:DCTD_def}), using a network architecture \emph{identical} to the baseline ATOM tracker. In particular, we employ the same bounding box parameterization as for object detection (Section~\ref{experiments:object_detection}) and sample $M=128$ boxes during training from a proposal distribution (\ref{eq:proposal_distribution}) generated by $L=2$ Gaussians with standard deviations $\sigma_1=0.05$, $\sigma_2=0.5$. During tracking, we follow the same procedure as in ATOM, sampling $10$ boxes in each frame followed by gradient ascent to refine the estimate generated by the classification module. The inference speed of our approach is thus identical to ATOM, running at over 30 FPS on a single NVIDIA GT-1080 GPU.

\begin{figure}[t]
    \centering
    \subfloat[]{
      \centering
            \begin{tikzpicture}[scale=0.639825]
                \pgfplotsset{
                    y axis style/.style={
                        yticklabel style=#1,
                        ylabel style=#1,
                        y axis line style=#1,
                        ytick style=#1
                  }
                }
            
                \begin{axis}[
                        axis y line*=left,
                        xlabel={Number of gradient ascent iterations $T$},
                        ylabel={AP (\%)},
                        xtick={0, 10, 20, 30, 40, 50, 60},
                        legend pos=north east,
                        grid style=dashed,
                        y tick label style={
                            /pgf/number format/.cd,
                                fixed,
                                fixed zerofill,
                                precision=1,
                            /tikz/.cd
                        },
                        every axis plot/.append style={thick},
                        y axis style=blue!65!black,
                    ]
                \addplot[smooth,mark=*,blue] 
                     plot [error bars/.cd, y dir = both, y explicit]
                     table[row sep=crcr, x index=0, y index=1]{
                    0 38.15\\
                    1 38.35\\
                    2 38.57\\
                    4 38.89\\
                    8 39.08\\
                    16 39.12\\
                    32 39.14\\
                    64 39.13\\
                    };
                \end{axis}
                
                \begin{axis}[
                    axis y line*=right,
                    axis x line=none,
                    xlabel={Number of gradient ascent iterations $T$},
                    ylabel={FPS},
                    legend pos=north east,
                    grid style=dashed,
                    y tick label style={
                        /pgf/number format/.cd,
                            fixed,
                            fixed zerofill,
                            precision=0,
                        /tikz/.cd
                    },
                    every axis plot/.append style={thick},
                    y axis style=red!65!black,
                ]
                ]
                \addplot[smooth,mark=*,red] 
                     plot [error bars/.cd, y dir = both, y explicit]
                     table[row sep=crcr, x index=0, y index=1]{
                    0 8.4\\
                    1 8\\
                    2 7.4\\
                    4 6.7\\
                    8 5.6\\
                    16 4.2\\
                    32 2.7\\
                    64 1.6\\
                    };
                \end{axis}
            \end{tikzpicture}%\vspace{-1mm}
      \label{fig:detection_impact_of_grad_iters}
    }
    \quad
    \subfloat[]{
        \centering
        \includegraphics[width=0.4375\textwidth]{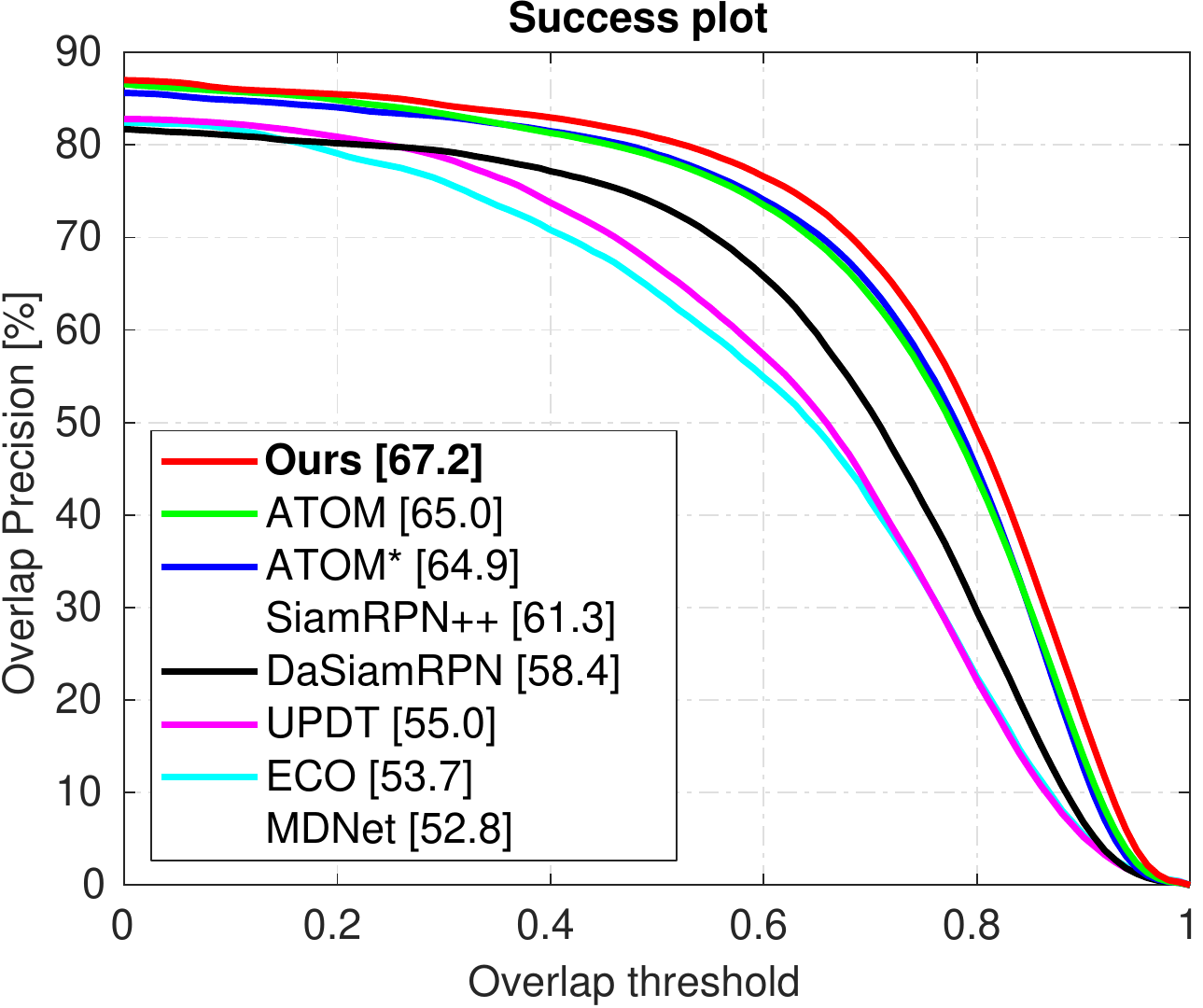}%\vspace{-1mm}
        \label{fig:uav}
    }
    \caption{\textbf{(a)} Impact of the number of gradient ascent iterations $T$ on performance (AP) and inference speed (FPS), for the object detection task on the \emph{2017 val} split of the COCO~\cite{lin2014microsoft} dataset. \textbf{(b)} Success plot on the UAV123~\cite{UAV123} visual tracking dataset, showing the overlap precision $\text{OP}_H$ as a function of the overlap threshold $H$.}
\end{figure}
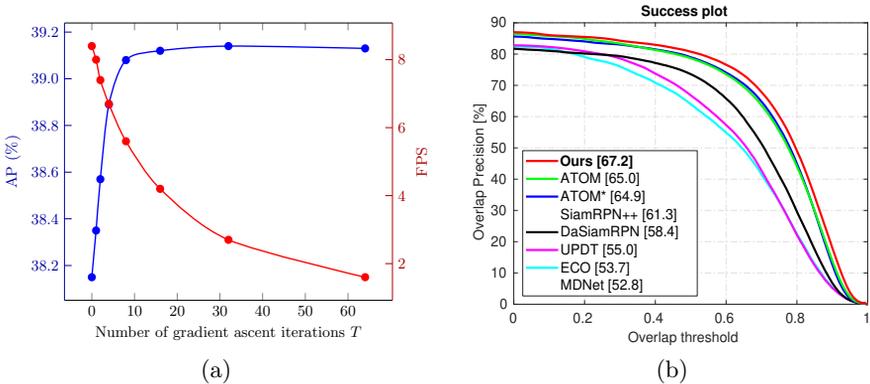

We demonstrate results on two standard tracking benchmarks: TrackingNet \cite{TrackingNet} and UAV123 \cite{UAV123}. TrackingNet contains challenging videos sampled from YouTube, with a test set of 511 videos. The main metric is the Success, defined as the average IoU overlap with the ground truth. UAV123 contains 123 videos captured from a UAV, and includes small and fast-moving objects. We report the overlap precision metric ($\text{OP}_{H}$), defined as the percentage of frames having bounding box IoU overlap larger than a threshold $H$. The final AUC score is computed as the average OP over all thresholds $H\in[0,1]$. Hyperparameters are set on the OTB \cite{OTB2015} and NFS \cite{NfS} datasets, containing 100 videos each. Due to the significant challenges imposed by the limited supervision and generic nature of the tracking problem, there are no competitive baselines employing direct bounding box regression. Current state-of-the-art employ either confidence-based regression, as in ATOM, or anchor-based bounding box regression techniques \cite{DaSiamRPN,SiamRPN++}. We therefore only compare with the ATOM baseline and include other recent state-of-the-art methods in the comparison. As in Section~\ref{experiments:object_detection}, we also compare with a version (denoted ATOM$^*$) of the IoU-Net-based ATOM employing the same training and inference settings as our final approach. The results are shown in Table~\ref{tab:tracking}, and the success plot on UAV123 is shown in Figure~\ref{fig:uav}. Our approach achieves a significant $2.5\%$ and $2.2\%$ absolute improvement over ATOM on the overall metric on TrackingNet and UAV123, respectively. Note that the improvements are most prominent for high-accuracy boxes, as indicated by OP$_{0.75}$. Our approach also outperforms the recent SiamRPN++ \cite{SiamRPN++}, which employs anchor-based bounding box regression \cite{Ren2015FasterRT,Redmon2016YOLO9000BF} and a much deeper backbone network (ResNet50) compared to ours (ResNet18). Figure~\ref{fig:intro_fig} (bottom) visualizes an illustrative example of the target density $p(y|x;\theta) \propto e^{f_\theta(x,y)}$ predicted by our approach during tracking. As illustrated, it predicts flexible densities which qualitatively capture meaningful uncertainty in challenging cases.

\begin{table*}[t]
\centering
\caption{Results for the age estimation task on the UTKFace~\cite{zhang2017age} dataset. Gradient-based refinement using our proposed method consistently improves MAE (lower is better) for the age predictions produced by a variety of different baselines.}
\resizebox{1.0\textwidth}{!}{%
\begin{tabular}{cccccccc}
\toprule
\multicolumn{1}{c}{+Refine} &\multicolumn{1}{c}{Niu et al. \cite{niu2016ordinal}} &\multicolumn{1}{c}{Cao et al. \cite{cao2019consistent}} &\multicolumn{1}{c}{Direct} &\multicolumn{1}{c}{Gaussian} &\multicolumn{1}{c}{Laplace} &\multicolumn{1}{c}{Softmax (CE, $L^2$)} &\multicolumn{1}{c}{Softmax (CE, $L^2$, Var)}\\ 
\midrule
 &5.74 $\pm$ 0.05 &5.47 $\pm$ 0.01 &4.81 $\pm$ 0.02 &4.79 $\pm$ 0.06 &4.85 $\pm$ 0.04 &4.78 $\pm$ 0.05 &4.81 $\pm$ 0.03\\
\checkmark &- &- &\textbf{4.65} $\pm$ 0.02 &4.66 $\pm$ 0.04 &4.81 $\pm$ 0.04 &\textbf{4.65} $\pm$ 0.04 &4.69 $\pm$ 0.03\\
\bottomrule
\end{tabular}}%\vspace{-4.0mm}
\label{table:age_estimation}%
\end{table*}

%%%%%%%%%%%%%%%%%%%%%%%%%%%%%%%%%%%%%%%%%%%%%%%%%%%%%%%%%%%%%%%%%%%%%%%%%%%%%%%
\subsection{Age Estimation}
\label{experiments:age_estimation}

To demonstrate the general applicability of our proposed method, we also perform experiments on regression tasks not involving bounding boxes. In age estimation, we are given a cropped image $x \in \mathbb{R}^{h \times w \times 3}$ of a person's face, and the task is to predict his/her age $y \in \mathbb{R}_+$. We utilize the UTKFace~\cite{zhang2017age} dataset, specifically the subset of $16\thinspace434$ images used by Cao et al.~\cite{cao2019consistent}. We also utilize the dataset split employed in \cite{cao2019consistent}, with $3\thinspace287$ test images and $11\thinspace503$ images for training. Additionally, we use $1\thinspace644$ of the training images for validation. Methods are evaluated in terms of the Mean Absolute Error (MAE). The DNN architecture $f_{\theta}(x,y)$ of our model first extracts ResNet50~\cite{he2016deep} features $g_x \in \mathbb{R}^{2048}$ from the input image $x$. The age $y$ is processed by four fully-connected layers, generating $g_y \in \mathbb{R}^{128}$. The two feature vectors are then concatenated and processed by two fully-connected layers, outputting $f_{\theta}(x, y) \in \mathbb{R}$. We apply our proposed method to refine the age predicted by baseline models, using the gradient ascent maximization of $f_{\theta}(x, y)$ detailed in Section~\ref{DCTDs:prediction}. All baseline DNN models employ a similar architecture, including an identical ResNet50 for feature extraction and the same number of fully-connected layers to output either the age $y \in \mathbb{R}$ (\textit{Direct}), mean and variance parameters for Gaussian and Laplace distributions, or to output logits for $C$ discretized classes (\textit{Softmax}). The results are found in Table~\ref{table:age_estimation}. We observe that age refinement provided by our method consistently improves the accuracy of the predictions generated by the baselines. For \textit{Direct}, e.g., this refinement marginally decreases inference speed from $49$ to $36$ FPS.

\begin{table*}[t]
\centering
\caption{Results for the head-pose estimation task on the BIWI~\cite{fanelli2013random} dataset. Gradient-based refinement using our proposed method consistently improves the average MAE (lower is better) for yaw, pitch and roll for the predicted pose produced by our baselines.}
\resizebox{1.0\textwidth}{!}{%
\begin{tabular}{cccccccc}
\toprule
\multicolumn{1}{c}{+Refine} &\multicolumn{1}{c}{Gu et al. \cite{gu2017dynamic}} &\multicolumn{1}{c}{Yang et al. \cite{yang2019fsa}} &\multicolumn{1}{c}{Direct} &\multicolumn{1}{c}{Gaussian} &\multicolumn{1}{c}{Laplace} &\multicolumn{1}{c}{Softmax (CE, $L^2$)} &\multicolumn{1}{c}{Softmax (CE, $L^2$, Var)}\\ 
\midrule
          &3.66 &3.60 &3.09 $\pm$ 0.07 &3.12 $\pm$ 0.08 &3.21 $\pm$ 0.06 &3.04 $\pm$ 0.08 &3.15 $\pm$ 0.07\\
\checkmark &- &- &3.07 $\pm$ 0.07 &3.11 $\pm$ 0.07 &3.19 $\pm$ 0.06 &\textbf{3.01} $\pm$ 0.07 &3.11 $\pm$ 0.06\\
\bottomrule
\end{tabular}}%\vspace{-4.0mm}
\label{table:head_pose_estimation}%
\end{table*}

%%%%%%%%%%%%%%%%%%%%%%%%%%%%%%%%%%%%%%%%%%%%%%%%%%%%%%%%%%%%%%%%%%%%%%%%%%%%%%%
\subsection{Head-Pose Estimation}
\label{experiments:head_pose_estimation}

Lastly, we evaluate our method on the task of head-pose estimation. In this case, we are given an image $x \in \mathbb{R}^{h \times w \times 3}$ of a person, and the aim is to predict the orientation $y \in \mathbb{R}^3$ of his/her head, where $y$ is the yaw, pitch and roll angles. We utilize the BIWI~\cite{fanelli2013random} dataset, specifically the processed dataset provided by Yang et al.~\cite{yang2019fsa}, in which the images have been cropped to faces detected using MTCNN~\cite{zhang2016joint}. We also employ protocol 2 as defined in \cite{yang2019fsa}, with $10\thinspace613$ images for training and $5\thinspace065$ images for testing. Additionally, we use $1\thinspace010$ training images for validation. The methods are evaluated in terms of the average MAE for yaw, pitch and roll. The network architecture of the DNN $f_{\theta}(x, y)$ defining our model takes the image $x \in \mathbb{R}^{h \times w \times 3}$ and orientation $y \in \mathbb{R}^3$ as inputs, but is otherwise identical to the age estimation case (Section~\ref{experiments:age_estimation}). Our model is again evaluated by applying the gradient-based refinement to the predicted orientation $y \in \mathbb{R}^3$ produced by a number of baseline models. We use the same baselines as for age estimation, and apart from minor changes required to increase the output dimension from $1$ to $3$, identical network architectures are also used. The results are found in Table~\ref{table:head_pose_estimation}, and also in this case we observe that refinement using our proposed method consistently improves upon the baselines.
\section{Conclusion}
\label{section: conclusion}

We proposed a general and conceptually simple regression method with a clear probabilistic interpretation. It models the conditional target density $p(y | x)$ by predicting the un-normalized density through a DNN $f_\theta(x, y)$, taking the input-target pair $(x,y)$ as input. This energy-based model $p(y | x; \theta) = e^{f_{\theta}(x, y)}/Z(x, \theta)$ of $p(y | x)$ is trained by directly minimizing the associated negative log-likelihood, employing Monte Carlo importance sampling to approximate the partition function $Z(x, \theta)$. At test time, targets are predicted by maximizing the DNN output $f_\theta(x, y)$ w.r.t.\ $y$ via gradient-based refinement. Extensive experiments performed on four diverse computer vision tasks demonstrate the high accuracy and wide applicability of our method. Future directions include exploring improved architectural designs, studying other regression applications, and investigating our proposed method's potential for aleatoric uncertainty estimation.

\parsection{Acknowledgments}
This research was supported by the Swedish Foundation for Strategic Research via \emph{ASSEMBLE}, the Swedish Research Council via \emph{Learning flexible models for nonlinear dynamics}, the ETH Z\"urich Fund (OK), a Huawei Technologies Oy (Finland) project, an Amazon AWS grant, and Nvidia.

\clearpage
\bibliographystyle{splncs04}
\bibliography{references}

\clearpage

\renewcommand{\thefigure}{S\arabic{figure}}
\setcounter{figure}{0}

\renewcommand{\thetable}{S\arabic{table}}
\setcounter{table}{0}

\renewcommand{\thealgorithm}{S\arabic{algorithm}}
\setcounter{algorithm}{0}

\renewcommand{\theequation}{S\arabic{equation}}
\setcounter{equation}{0}

\makeatletter
\newcommand{\@chapapp}{\relax}%
\makeatother

\section*{\centering{Supplementary Material}}

In this supplementary material, we provide additional details and results. It consists of \ref{appendix:prediction} - \ref{appendix:head_pose_estimation}. \ref{appendix:prediction} contains a detailed algorithm for our employed prediction procedure. \ref{appendix:illustrative_example} contains details on the illustrative 1D regression problem in Figure~2 in the main paper. Further details on the employed training and inference procedures are provided in \ref{appendix:object_detection} for the object detection experiments, and in \ref{appendix:visual_tracking} for the visual tracking experiments. Lastly, \ref{appendix:age_estimation} and \ref{appendix:head_pose_estimation} contain details and full results for the experiments on age estimation and head-pose estimation, respectively. Note that equations, tables, figures and algorithms in this supplementary document are numbered with the prefix "S". Numbers without this prefix refer to the main paper.

\appendix
\begin{appendices}
\renewcommand{\thesection}{\appendixname~\Alph{section}}
\renewcommand{\thesubsection}{\Alph{section}.\arabic{subsection}}
\section{Prediction Algorithm}
\label{appendix:prediction}
Our prediction procedure (Section 3.3) is detailed in Algorithm~\ref{algo:prediction}, where $\lambda$ denotes the gradient ascent step-length, $\eta$ is a decay of the step-length and $T$ is the number of iterations. In our experiments, we fix $T$ (typically, $T = 10$) and select $\{\lambda, \eta\}$ using grid search on a validation set.

\begin{algorithm}
\caption{Prediction via gradient-based refinement.}
\label{algo:prediction}
\textbf{Input:} $x^\star$, $\hat{y}$, $T$, $\lambda$, $\eta$.
\begin{algorithmic}[1]
    \State $y \gets \hat{y}$.
    \For{\texttt{$t = 1, \dots, T$}}
        \State \texttt{PrevValue} $\gets$ $f_{\theta}(x^\star, y)$.
        \State $\Tilde{y} \gets y + \lambda \nabla_{y} f_{\theta}(x^\star, y)$.
        \State \texttt{NewValue} $\gets$ $f_{\theta}(x^\star, \Tilde{y})$.
        \If { $\texttt{NewValue} > \texttt{PrevValue}$}
            \State $y \gets \Tilde{y}$.
        \Else
            \State $\lambda \gets \eta \lambda$.
        \EndIf
    \EndFor
    \State \textbf{Return} $y$.
\end{algorithmic}
\end{algorithm}\vspace{-3mm}
\section{Illustrative Example}
\label{appendix:illustrative_example}

The ground truth conditional target density $p(y | x)$ in Figure~2 is defined by a mixture of two Gaussian components (with weights $0.2$ and $0.8$) for $x < 0$, and a log-normal distribution (with $\mu = 0.0$, $\sigma= 0.25$) for $x \geq 0$. The training data $\{(x_i, y_i)\}_{i=1}^{2000}$ was generated by uniform random sampling of $x$, $x_i \sim U(-3, 3)$. Both models were trained for $75$ epochs with a batch size of $32$ using the ADAM~\cite{kingma2014adam} optimizer.

The Gaussian model is defined using a DNN $f_\theta(x)$ according to,
\begin{equation}
\begin{gathered}
    p(y | x; \theta) = \mathcal{N}\big(y; \mu_{\theta}(x),\,\sigma^2_{\theta}(x)\big),\\
    f_{\theta}(x) = \rvect{\mu_{\theta}(x), \log\sigma^2_{\theta}(x)}^{\Transp} \in \mathbb{R}^2.
\end{gathered}
\end{equation}
It is trained by minimizing the negative log-likelihood, corresponding to the loss,
\begin{equation}
    J(\theta) = \frac{1}{n} \sum_{i=1}^{n} \frac{(y_i - \mu_\theta(x_i))^2}{\sigma_\theta^{2}(x_i)} + \log \sigma_\theta^{2}(x_i).
\end{equation}
The DNN $f_\theta$ is a simple feed-forward neural network, containing two shared fully-connected layers (dimensions: $1 \rightarrow 10$, $10 \rightarrow 10$) and two identical heads for $\mu$ and $\log\sigma^2$ of three fully-connected layers ($10 \rightarrow 10$, $10 \rightarrow 10$, $10 \rightarrow 1$).

Our proposed model $p(y | x; \theta) = e^{f_{\theta}(x, y)}/Z(x, \theta)$ (Eq.~1 in the paper) is defined using a feed-forward neural network $f_\theta(x, y)$ containing two fully-connected layers ($1 \rightarrow 10$, $10 \rightarrow 10$) for both $x$ and $y$, and three fully-connected layers ($20 \rightarrow 10$, $10 \rightarrow 10$, $10 \rightarrow 1$) processing the concatenated $(x, y)$ feature vector. It is trained using $M = 1024$ samples from a proposal distribution $q(y | y_i)$ (Eq.~5 in the paper) with $L=2$ and variances $\sigma_1^2 = 0.1^2$, $\sigma_2^2 = 0.8^2$.
\section{Object Detection}
\label{appendix:object_detection}

Here, we provide further details about the network architectures, training procedure, and hyperparameters used for our experiments on object detection (Section~4.1 in the paper).

\subsection{Network Architecture}
\begin{figure*}[t]
	\centering
	\subfloat[Faster-RCNN.]{
		\centering
		\includegraphics[trim = 0 0 0 0, width=0.315\textwidth]{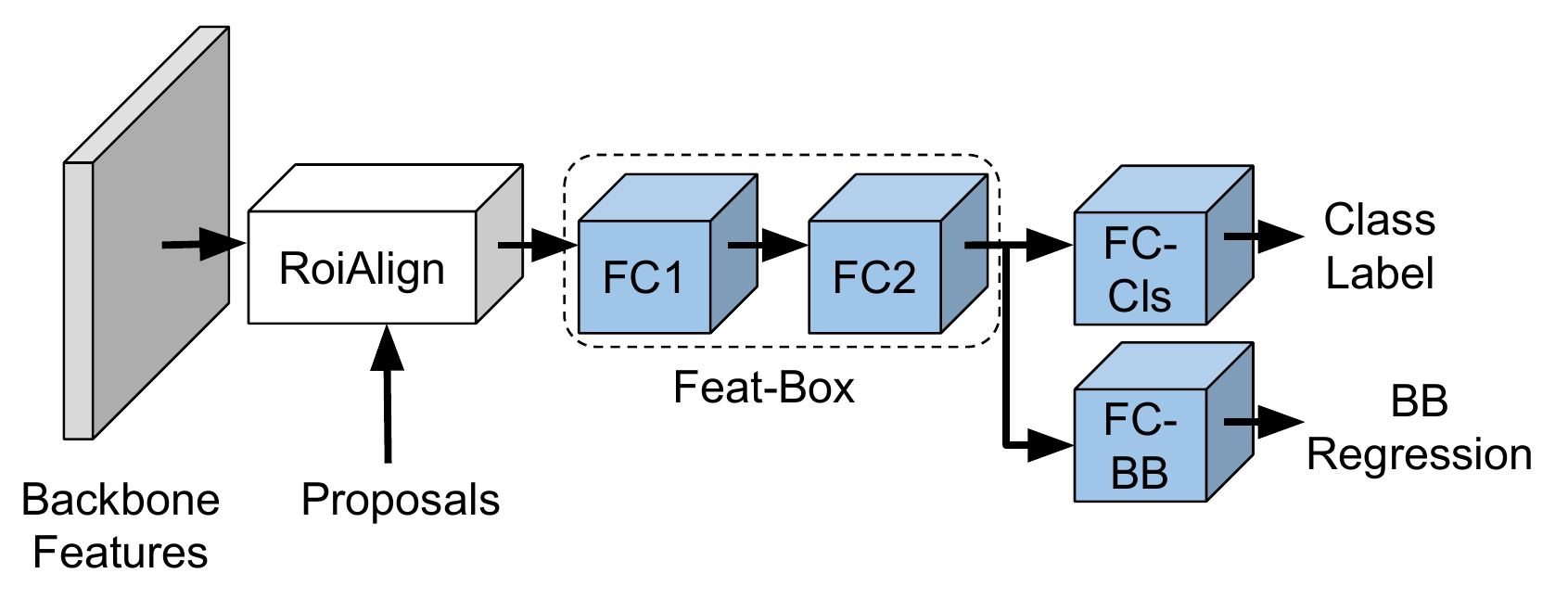}
		\label{fig:frcnn}
	}%
	%\quad
	\subfloat[Laplace/Gaussian.]{
		\centering
		\includegraphics[trim = 0 0 0 0, width=0.315\textwidth]{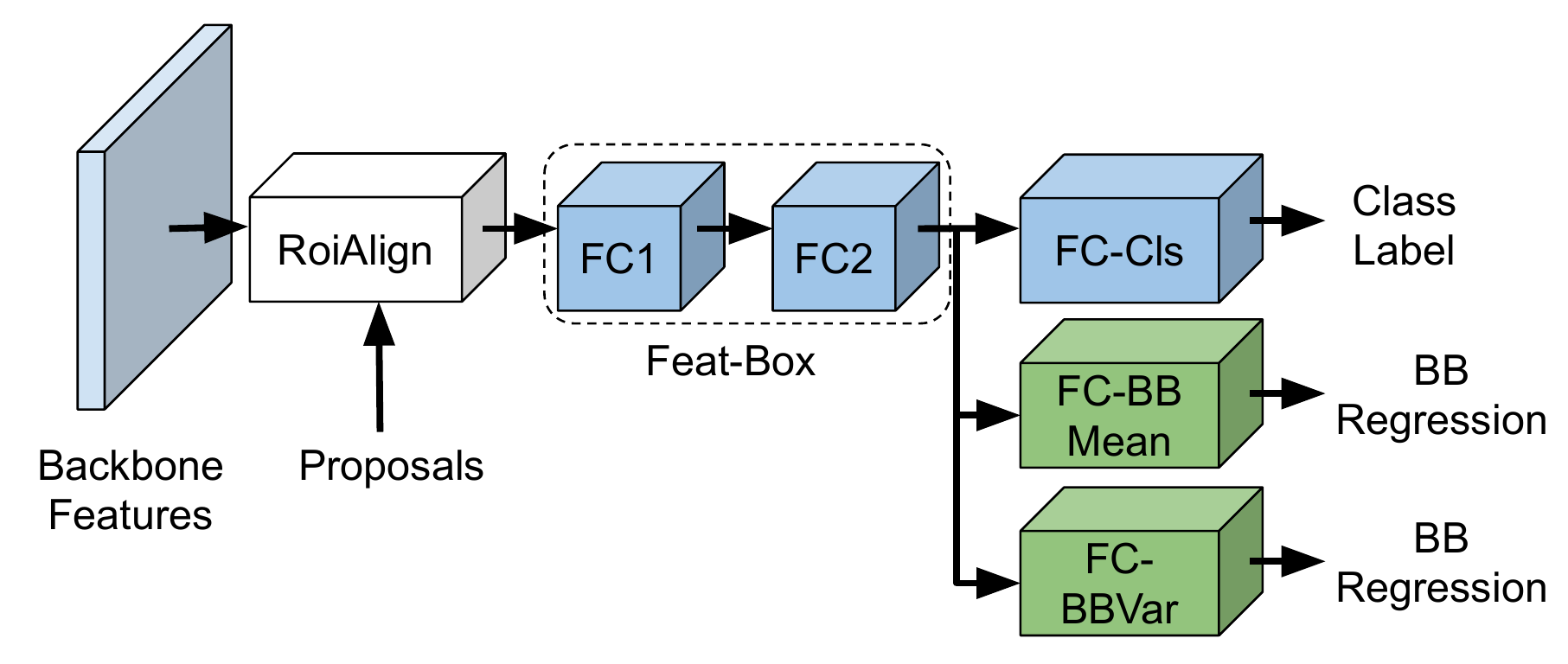}
		\label{fig:gauss-frcnn}
	}%
	%\quad
	\subfloat[IoU-Net/Ours.]{
		\centering
		\includegraphics[trim = 0 0 0 0, width=0.315\textwidth]{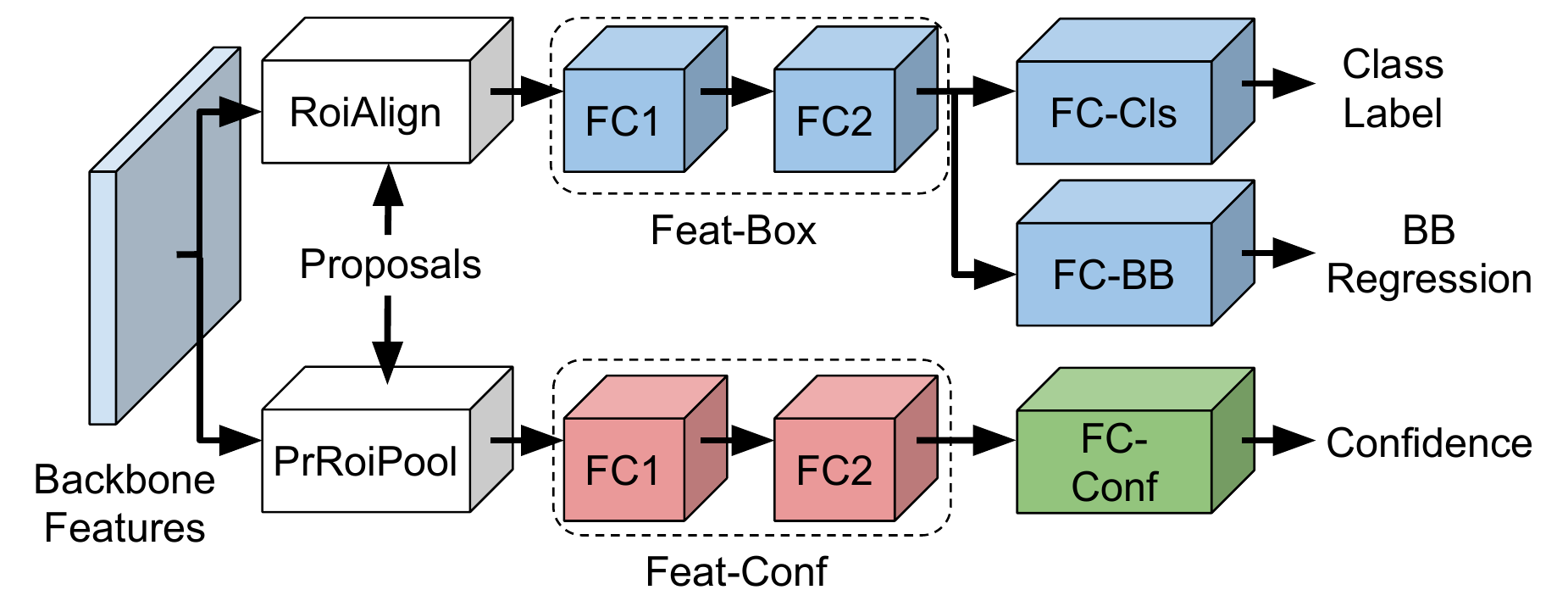}
		\label{fig:ours-frcnn}
	}
	\caption{Network architectures for the different object detection networks used in our experiments (Section~4.1 in the paper). The backbone feature extractor (ResNet50-FPN), and the region proposal network (RPN) is not shown for clarity. We do not train the blocks in blue color, using the pre-trained Faster-RCNN weights from \cite{massa2018mrcnn} instead. The blocks in red are initialized with the pre-trained Faster-RCNN weights and fine-tuned. The blocks in green on the other hand are trained from scratch.}
	\label{fig:detectionarch}
\end{figure*}

We use the Faster-RCNN~\cite{Ren2015FasterRT} detector with ResNet50-FPN~\cite{lin2017feature} as our baseline. As visualized in Figure~\ref{fig:frcnn}, Faster-RCNN generates object proposals using a region proposal network (RPN). The features from the proposal regions are then pooled to a fixed-sized feature map using the RoiPool layer~\cite{Girshick2015FastR}. The pooled features are then passed through a feature extractor (denoted Feat-Box) consisting of two fully-connected (FC) layers. The output feature vector is then passed through two parallel FC layers, one which predicts the class label (denoted FC-Cls), and another which regresses the offsets between the proposal and the ground truth box (denoted FC-BB). We use the PyTorch implementation for Faster-RCNN from~\cite{massa2018mrcnn}. Note that we use the RoiAlign \cite{He2017MaskR} layer instead of RoiPool in our experiments as it has been shown to achieve better performance~\cite{He2017MaskR}.

For the Gaussian and Laplace probabilistic models (Gaussian and Laplace in Table~2 in the paper), we replace the FC-BB layer in Faster-RCNN with two parallel FC layers, denoted FC-BBMean and FC-BBVar, which predict the mean and the log-variance of the distribution modeling the offset between the proposal and the ground truth box for each coordinate. This architecture is shown in Figure~\ref{fig:gauss-frcnn}. For the mixtures of $K = \{2, 4, 8\}$ Gaussians, we duplicate FC-BBMean and FC-BBVar $K$ times, and add an FC layer for predicting the $K$ component weights. For the cVAE, FC-BBMean and FC-BBVar instead outputs the mean and log-variance of a Gaussian distribution for the latent variable $z \in \mathbb{R}^{10}$. Duplicates of FC-BBMean and FC-BBVar, modified to also take sampled $z$ values as input, then predicts the mean and log-variance of the distribution modeling the bounding box offset.

% \begin{figure*}[t]
% 	\centering
% 	\newcommand{\wid}{1.0\textwidth}%
% 	\begin{subfigure}[t]{0.33\textwidth}
% 		\centering
% 		\includegraphics[trim = 0 0 0 0, width = \wid]{figures/network_frcnn}
% 		\caption{Faster-RCNN.}\label{fig:frcnn}
% 	\end{subfigure}%
% 	\begin{subfigure}[t]{0.33\textwidth}
% 		\centering
% 		\includegraphics[trim = 0 0 0 0, width = \wid]{figures/network_prob}
% 		\caption{Laplace/Gaussian.}\label{fig:gauss-frcnn}
% 	\end{subfigure}%
% 	\begin{subfigure}[t]{0.33\textwidth}
% 		\centering
% 		\includegraphics[trim = 0 0 0 0, width = \wid]{figures/network_score}
% 		\caption{IoU-Net/Ours.}\label{fig:ours-frcnn}
% 	\end{subfigure}
% 	\caption{Network architectures for the different object detection networks used in our experiments (Section~4.1 in the paper). The backbone feature extractor (ResNet50-FPN), and the region proposal network (RPN) is not shown for clarity. We do not train the blocks in blue color, using the pre-trained Faster-RCNN weights from \cite{massa2018mrcnn} instead. The blocks in red are initialized with the pre-trained Faster-RCNN weights and fine-tuned. The blocks in green on the other hand are trained from scratch.}
% 	\label{fig:detectionarch}
% \end{figure*}

For our confidence-based IoU-Net \cite{jiang2018acquisition} models (IoU-Net and IoU-Net$^*$ in Table~2), we use the same network architecture as employed in the original paper, shown in Figure~\ref{fig:ours-frcnn}. That is, we add an additional branch to predict the IoU overlap between the proposal box and the ground truth. This branch uses the PrRoiPool \cite{jiang2018acquisition} layer to pool the features from the proposal regions. The pooled features are passed through a feature extractor (denoted Feat-Conf) consisting of two FC layers. The output feature vector is passed through another FC layer, FC-Conf, which predicts the IoU. We use an identical architecture for our approach, but train it to output $f_\theta(x,y)$ in $p(y | x; \theta) = e^{f_{\theta}(x, y)}/Z(x, \theta)$ instead.

\subsection{Training}
We use the pre-trained weights for Faster-RCNN from~\cite{massa2018mrcnn}. Note that the bounding box regression in Faster-RCNN is trained using a direct method, with an Huber loss~\cite{huber1964robust}. We trained the other networks in Table~2 in the paper (Gaussian, Gaussian Mixt. 2, Gaussian Mixt. 4, Gaussian Mixt. 8, Gaussian cVAE, Laplace, IoU-Net, IoU-Net$^*$ and Ours) on the MS-COCO \cite{lin2014microsoft} training split (\emph{2017 train}) using stochastic gradient descent (SGD) with a batch size of 16 for 60k iterations. The base learning rate $lr_\text{base}$ is reduced by a factor of $10$ after 40k and 50k iterations, for all the networks. We also warm up the training by linearly increasing the learning rate from $\frac{1}{3}lr_\text{base}$ to $lr_\text{base}$ during the first 500 iterations. We use a weight decay of $0.0001$ and a momentum of $0.9$. For all the networks, we only trained the newly added layers, while keeping the backbone and the region proposal network fixed. 

For the Gaussian, mixture of Gaussians, cVAE and Laplace models, we only train the final predictors (FC-BBMean and FC-BBVar), while keeping the class predictor (FC-Cls) and the box feature extractor (Feat-Box) fixed. We also tried fine-tuning the FC-Cls and Feat-Box weights for the Gaussian model, with different learning rate settings, but obtained worse performance on the validation set. The weights for both FC-BBMean and FC-BBVar were initialized with zero mean Gaussian with standard deviation of $0.001$. All these models were trained with a base learning rate $lr_\text{base} = 0.005$ by minimizing the negative log-likelihood, except for the cVAE which is trained by maximizing the ELBO (using $128$ sampled $z$ values to approximate the expectation). 

For the IoU-Net, IoU-Net$^*$ and our proposed model, we only trained the newly added confidence branch. We found it beneficial to initialize the feature extractor block (Feat-Conf) with the corresponding weights from Faster-RCNN, i.e.\ the Feat-Box block. The weights for the predictor FC-Conf were initialized with zero mean Gaussian with standard deviation of $0.001$. As in the original paper~\cite{jiang2018acquisition}, we used a base learning rate $lr_\text{base} = 0.01$ for the IoU-Net and IoU-Net$^*$ networks. For our proposed model, we used $lr_\text{base} = 0.001$ due to the different scaling of the loss. Note that we did not perform any parameter tuning for setting the learning rates. We generate $128$ proposals for each ground truth box during training. For the IoU-Net, we use the proposal generation strategy mentioned in the original paper~\cite{jiang2018acquisition}. That is, for each ground truth box, we generate a large set of candidate boxes which have an IoU overlap of at least $0.5$ with the ground truth, and uniformly sample $128$ proposals from this candidate set w.r.t.\ the IoU. For IoU-Net$^*$ and our model, we sample boxes from a proposal distribution (Eq.~5 in the paper) generated by $L=3$ Gaussians with standard deviations of $0.0375$, $0.075$ and $0.15$. The IoU-Net and IoU-Net$^*$ are trained by minimizing the Huber loss between the predicted IoU and the ground truth, while our model is trained by minimizing the negative log likelihood of the training data (Eq.~4 in the paper).

\begin{table}[t]
\begin{center}
\caption{Impact of $L$ and $\{\sigma_{l}\}_{l=1}^{L}$ in the proposal distribution $q(y|y_i)$~(Eq.~5 in the paper), for the object detection task on the \emph{2017 val} split of COCO~\cite{lin2014microsoft}.}
\begin{tabular}{|@{\hspace{0.5cm}}c@{\hspace{0.5cm}}|l@{\hspace{1cm}}|l@{\hspace{1cm}}|}
\multicolumn{1}{c}{$L$} &\multicolumn{1}{c}{$\{\sigma_{l}\}_{l=1}^{L}$} &\multicolumn{1}{c}{AP (\%)}\\ 
\hline
1 &\{0.01875\} &38.07\\
1 &\{0.0375\} &38.47\\
1 &\{0.075\} &37.52\\
1 &\{0.15\} &35.05\\
\hline
\hline
2 &\{0.025, 0.1\} &38.97\\
2 &\{0.0375, 0.15\} &39.05\\
2 &\{0.04375, 0.175\} &39.07\\
2 &\{0.05, 0.2\} &39.02\\
\hline
2 &\{0.0125, 0.025\} &38.19\\
2 &\{0.025, 0.05\} &38.65\\
2 &\{0.075, 0.15\} &37.14\\
\hline
\hline
3 &\{0.0125, 0.025, 0.05\} &38.61\\
3 &\{0.025, 0.05, 0.1\} &38.95\\
3 &\{0.0375, 0.075, 0.15\} &\textbf{39.11}\\
3 &\{0.04375, 0.0875, 0.175\} &39.00\\
3 &\{0.05, 0.1, 0.2\} &38.76\\
3 &\{0.0625, 0.125, 0.25\} &37.96\\
3 &\{0.075, 0.15, 0.3\} &37.42\\
\hline
\end{tabular}\vspace{-3mm}
\label{table:detection_ablation_full}%
\end{center}
\end{table}

\subsection{Analysis of Proposal Distribution}
An extensive ablation study for the number of components $L$ and standard deviations $\{\sigma_{l}\}_{l=1}^{L}$ in the proposal distribution $q(y|y_i) = \frac{1}{L} \sum_{l=1}^{L} \mathcal{N}(y; y_i, \sigma_{l}^{2})$ (Eq.~5 in the paper) is provided in Table~\ref{table:detection_ablation_full}, which is an extended version of Table~1 in the paper. We find that $L=1$ downgrades performance, while there is no significant difference between $L=2$ and $L=3$. For $L \in \{2, 3\}$, the results are not particularly sensitive to the specific choice of $\{\sigma_{l}\}_{l=1}^{L}$, but benefits from including \emph{both} small and relatively large values in $\{\sigma_{l}\}_{l=1}^{L}$.

\subsection{Inference}
The inference in both the Gaussian and Laplace models is identical to the one employed by Faster-RCNN, except the output mean is taken as the prediction. Thus, we do not utilize the output variances during inference. For the mixture of $K = \{2, 4, 8\}$ Gaussians, we compute the mean of the distribution and take that as our prediction. Instead picking the component with the largest weight and taking its mean as the prediction resulted in somewhat worse validation performance. For cVAE, we approximately compute the mean (using $128$ samples) of the distribution and take that as our prediction.

For IoU-Net and IoU-Net$^*$, we perform IoU-Guided NMS as described in \cite{jiang2018acquisition}, followed by gradient-based refinement (Algorithm~\ref{algo:prediction}). For our proposed approach we adopt the same NMS technique, but guide it with the values $f_\theta(x,y)$ predicted by our network instead. We use a step-length $\lambda = 0.5$ and step-length decay $\eta = 0.1$ for IoU-Net. For IoU-Net$^*$ and our approach we perform the gradient-based refinement in the relative bounding box parametrization $y = (c_x/w_0, c_y/h_0, \log w, \log h)$ (see Section~4.1 in the paper). Here, we employ different step-lengths for position and size. For IoU-Net$^*$, we use $\lambda = 0.002$ and $\lambda = 0.008$ respectively, with a decay of $\eta = 0.2$. For our proposed approach, we use $\lambda = 0.0001$ and  $\lambda =0.0004$ with $\eta = 0.5$. For all methods, these hyperparameters ($\lambda$ and $\eta$) were set using a grid search on the COCO validation split (\emph{2017 val}). We used $T=10$ refinement iterations for each of the three models. Note that since a given image $x$ can have multiple ground truth instances, multiple bounding boxes are usually refined. The gradient-based refinement then moves each individual box $y$ towards the maximum of a \emph{local} mode in $f_{\theta}(x, y)$. Thus, they will not converge to a single solution. Also note that $f_{\theta}(x, y)$ is class-conditional (as in the IoU-Net baseline), eliminating the risk of confusing neighboring objects of different classes.
\section{Visual Tracking}
\label{appendix:visual_tracking}

Here, we provide further details about the training procedure and hyperparameters used for our experiments on visual object tracking (Section~4.2 in the paper). 

\begin{table}[t]
\begin{center}
\caption{Impact of $L$ and $\{\sigma_{l}\}_{l=1}^{L}$ in the proposal distribution $q(y|y_i)$~(Eq.~5 in the paper), for the visual tracking task on the combined OTB~\cite{OTB2015} and NFS~\cite{NfS} datasets.}
\begin{tabular}{|@{\hspace{0.5cm}}c@{\hspace{0.5cm}}|l@{\hspace{1.25cm}}|l@{\hspace{1.25cm}}|l@{\hspace{1.25cm}}|l@{\hspace{1.25cm}}|}
\multicolumn{1}{c}{$L$} &\multicolumn{1}{c}{$\{\sigma_{l}\}_{l=1}^{L}$} &\multicolumn{1}{c}{OP$_{0.50}$ (\%)} &\multicolumn{1}{c}{OP$_{0.75}$ (\%)} &\multicolumn{1}{c}{AUC (\%)}\\ 
\hline
1 &\{0.05\} &75.77 &45.72 &63.37\\
\hline
2 &\{0.01, 0.1\} &77.25 &46.09 &61.48\\
2 &\{0.03, 0.3\} &79.27 &48.59 &63.65\\
2 &\{0.05, 0.5\} &\textbf{79.90} &\textbf{48.71} &\textbf{64.10}\\
2 &\{0.07, 0.7\} &78.41 &47.72 &62.75\\
\hline
\end{tabular}\vspace{-3mm}
\label{table:tracking_ablation}%
\end{center}
\end{table}

\subsection{Training}
We adopt the ATOM \cite{danelljan2019atom} tracker as our baseline, and use the PyTorch implementation and pre-trained weights from~\cite{pytracking}. ATOM trains an IoU-Net-based module to predict the IoU overlap between a candidate box and the ground truth, conditioned on the first-frame target appearance. The IoU predictor is trained by generating $16$ candidates for each ground truth box. The candidates are generated by adding a Gaussian noise for each ground truth box coordinate, while ensuring a minimum IoU overlap of $0.1$ between the candidate box and the ground truth. The network is trained by minimizing the squared error ($L^2$ loss) between the predicted and ground truth IoU.

Our proposed model is instead trained by sampling $128$ candidate boxes from a proposal distribution (Eq.~5 in the paper) generated by $L=2$ Gaussians with standard deviations of $0.05$ and $0.5$, and minimizing the negative log likelihood of the training data. An ablation study for the proposal distribution is found in Table~\ref{table:tracking_ablation}. We use the training splits of the TrackingNet \cite{TrackingNet}, LaSOT \cite{fan2019lasot}, GOT10k~\cite{huang2019got}, and COCO datasets for our training. Our network is trained for $50$ epochs, using the ADAM optimizer with a base learning rate of $0.001$ which is reduced by a factor of $5$ after every $15$ epochs. The rest of the training parameters are exactly the same is in ATOM. The ATOM$^*$ model is trained by using the exact same proposal distribution, datasets and settings. It only differs by the loss, which is the same squared error between the predicted and ground truth IoU as in the original ATOM.

\subsection{Inference}
During tracking, the ATOM tracker first applies the classification head network, which is trained online, to coarsely localize the target object. 10 random boxes are then sampled around this prediction, to be refined by the IoU prediction network. We only alter the final bounding box refinement step of the 10 given random initial boxes, and preserve all other settings as in the original ATOM tracker. The original version performs $T=5$ gradient ascent iterations with a step length of $\lambda = 1.0$. For our proposed model and the ATOM$^*$ version, we use $T=10$ iterations, employing the bounding box parameterization described in Section~4.1. For our approach, we set the step length to $\lambda=2 \cdot 10^{-4}$ for position and $\lambda= 10^{-3}$ for size dimensions. For ATOM$^*$, we use $\lambda=10^{-2}$ for position and $\lambda=5 \cdot  10^{-2}$ for size dimensions. These parameters were set on the separate validation set. For simplicity, we adopt the vanilla gradient ascent strategy employed in ATOM for the two other methods as well. That is, we have no decay ($\eta = 1$) and do not perform checks whether the confidence score is increasing in each iteration.

\subsection{Qualitative Results}

\begin{figure*}[!t]
	\centering%
	\newcommand{\wid}{0.25\textwidth}%
	\includegraphics*[trim = 50 70 50 73, width = \wid]{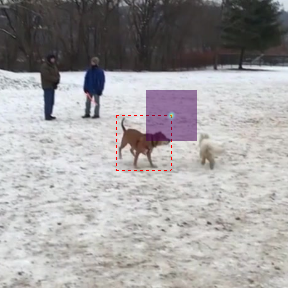}%
	\includegraphics*[trim = 50 70 50 73, width = \wid]{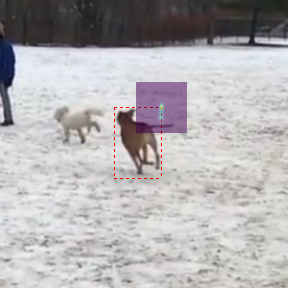}%
	\includegraphics*[trim = 50 70 50 73, width = \wid]{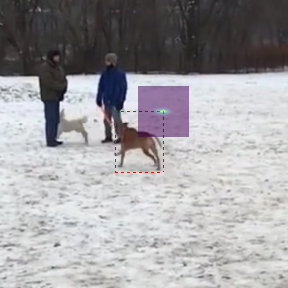}%
    \includegraphics*[trim = 50 70 50 73, width = \wid]{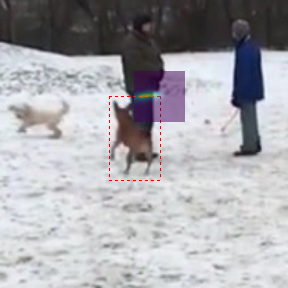}%\vspace{-2mm}
	\caption{Visualization of the conditional target density $p(y|x;\theta) \propto e^{f_\theta(x,y)}$ predicted by our network for the task of bounding box estimation in visual tracking. Since the target space $y\in \mathbb{R}^4$ is 4-dimensional, we visualize the density for different locations of the top-right corner as a heatmap, while the bottom-left is kept fixed at the tracker output (red box). Our network predicts flexible densities which qualitatively capture meaningful uncertainty in challenging cases.}%\vspace{-5mm}%
	\label{fig:qual_fig}%
\end{figure*}

Illustrative examples of the target density $p(y|x;\theta) \propto e^{f_\theta(x,y)}$ predicted by our approach during tracking are visualized in Figure~\ref{fig:qual_fig}.

% Figure~\ref{fig:intro_fig} (bottom) visualizes an illustrative example of the target density $p(y|x;\theta) \propto e^{f_\theta(x,y)}$ predicted by our approach during tracking. As illustrated, it predicts flexible densities which qualitatively capture meaningful uncertainty in challenging cases.

\section{Age Estimation}
\label{appendix:age_estimation}

In this appendix, further details on the age estimation experiments (Section~4.3 in the paper) are provided.

\subsection{Network Architecture}
The DNN architecture $f_{\theta}(x,y)$ of our proposed model first extracts ResNet50 features $g_x \in \mathbb{R}^{2048}$ from the input image $x$. The age $y$ is processed by four fully-connected layers (dimensions: $1 \rightarrow 16$, $16 \rightarrow 32$, $32 \rightarrow 64$, $64 \rightarrow 128$), generating $g_y \in \mathbb{R}^{128}$. The two feature vectors $g_x$, $g_y$ are then concatenated to form $g_{x,y} \in \mathbb{R}^{2048+128}$, which is processed by two fully-connected layers ($2048+128 \rightarrow 2048$, $2048 \rightarrow 1$), outputting $f_{\theta}(x, y) \in \mathbb{R}$.

\begin{table}[t]
\begin{center}
\caption{Impact of $\{\sigma_{l}\}_{l=1}^{2}$ in the proposal distribution $q(y|y_i)$~(Eq.~5 in the paper), for the age estimation task on our validation split of the UTKFace~\cite{zhang2017age} dataset.}
\begin{tabular}{|l@{\hspace{1cm}}|l@{\hspace{1cm}}|}
\multicolumn{1}{c}{$\{\sigma_{l}\}_{l=1}^{2}$} &\multicolumn{1}{c}{MAE}\\ 
\hline
\{0.1, 10\} &7.62\\
\{0.1, 20\} &\textbf{5.12}\\
\{0.01, 20\} &5.36\\
\{0.1, 40\} &5.24\\
\hline
\end{tabular}\vspace{-3mm}
\label{table:age_ablation}%
\end{center}
\end{table}

\subsection{Training}
Our model is trained using $M = 1024$ samples from a proposal distribution $q(y | y_i)$ (Eq.~5 in the paper) with $L=2$ and variances $\sigma_1^2 = 0.1^2$, $\sigma_2^2 = 20^2$. An ablation study for the variances is found in Table~\ref{table:age_ablation}. The model is trained for $75$ epochs with a batch size of $32$, using the ADAM optimizer with weight decay of $0.001$. The images $x$ are of size $200 \times 200$. For data augmentation, we use random flipping along the vertical axis and random scaling in the range $[0.7, 1.4]$. After random flipping and scaling, a random image crop of size $200 \times 200$ is also selected. The ResNet50 is imported from \texttt{torchvision.models} in PyTorch with the pretrained option set to true, all other network parameters are randomly initialized using the default initializer in PyTorch.

\subsection{Prediction}
For this experiment, we use a slight variation of Algorithm~\ref{algo:prediction}, which is found in Algorithm~\ref{algo:prediction_age}. There, $T$ is the number of gradient ascent iterations, $\lambda$ is the stepsize, $\Omega_1$ is an early-stopping threshold and $\Omega_2$ is a degeneration tolerance. Following IoU-Net, we set $T = 5$, $\Omega_1 = 0.001$ and $\Omega_2 = -0.01$. Based on the validation set, we select $\lambda = 3$. We refine a single estimate $\hat{y}$, predicted by each baseline model.

\begin{algorithm*}
\caption{Prediction via gradient-based refinement (variation).}\label{algo:prediction_age}
\textbf{Input:} $x^\star$, $\hat{y}$, $T$, $\lambda$, $\Omega_1$, $\Omega_2$. %\\
% \textbf{Output:} $y^\star$.
\begin{algorithmic}[1]
    \State $y \gets \hat{y}$.
    \For{\texttt{$t = 1, \dots, T$}}
        \State \texttt{PrevValue} $\gets$ $f_{\theta}(x^\star, y)$.
        \State $y \gets y + \lambda \nabla_{y} f_{\theta}(x^\star, y)$.
        \State \texttt{NewValue} $\gets$ $f_{\theta}(x^\star, y)$.
        
        \If {$|$\texttt{PrevValue} $-$ \texttt{NewValue}$| < \Omega_1$ \Or \hspace{0.5mm} $($\texttt{NewValue} $-$ \texttt{PrevValue}$) < \Omega_2$}
        \State \textbf{Return} $y$.
        \EndIf

    \EndFor
    \State \textbf{Return} $y$.
\end{algorithmic}
\end{algorithm*}

\subsection{Baselines}
All baselines are trained for $75$ epochs with a batch size of $32$, using the ADAM optimizer with weight decay of $0.001$. Identical data augmentation and parameter initialization as for our proposed model is used.

\parsection{Direct}
The DNN architecture of \textit{Direct} first extracts ResNet50 features $g_x \in \mathbb{R}^{2048}$ from the input image $x$. The feature vector $g_x$ is then processed by two fully-connected layers ($2048 \rightarrow 2048$, $2048 \rightarrow 1$), outputting the prediction $\hat{y} \in \mathbb{R}$. It is trained by minimizing either the Huber or $L^2$ loss.

\parsection{Gaussian}
The Gaussian model is defined using a DNN $f_\theta(x)$ according to,
\begin{equation}
\begin{gathered}
    p(y | x; \theta) = \mathcal{N}\big(y; \mu_{\theta}(x),\,\sigma^2_{\theta}(x)\big),\\
    f_{\theta}(x) = \rvect{\mu_{\theta}(x), \log\sigma^2_{\theta}(x)}^{\Transp} \in \mathbb{R}^2.
\end{gathered}
\end{equation}
It is trained by minimizing the negative log-likelihood, corresponding to the loss,
\begin{equation}
    J(\theta) = \frac{1}{n} \sum_{i=1}^{n} \frac{(y_i - \mu_\theta(x_i))^2}{\sigma_\theta^{2}(x_i)} + \log \sigma_\theta^{2}(x_i).
\end{equation}
The DNN architecture of $f_\theta(x)$ first extracts ResNet50 features $g_x \in \mathbb{R}^{2048}$ from the input image $x$. The feature vector $g_x$ is then processed by two heads of two fully-connected layers ($2048 \rightarrow 2048$, $2048 \rightarrow 1$) to output $\mu_{\theta}(x)$ and $\log\sigma^2_{\theta}(x)$. The mean $\mu_{\theta}(x)$ is taken as the prediction $\hat{y}$.

\begin{table}[t]
\begin{center}
\caption{Full results for the age estimation experiments. Gradient-based refinement using our proposed method consistently improves MAE (lower is better) for the age predictions outputted by a number of baselines.}
\begin{tabular}{|l||l|}
\multicolumn{1}{c}{Method} &\multicolumn{1}{c}{MAE}\\ 
\hline
Niu et al. \cite{niu2016ordinal}            &5.74 $\pm$ 0.05\\
Cao et al. \cite{cao2019consistent}      &5.47 $\pm$ 0.01 \\

\hline

% Direct regression (smoothL1)                        &4.79597 $\pm$ 0.0641251 \\
% Direct regression (smoothL1) + refinement           &4.744 $\pm$ 0.0615543 \\
% \hline

% Direct regression (L2)                        &4.80829 $\pm$ 0.0217812 \\
% Direct regression (L2) + refinement           &4.65449 $\pm$ 0.0240098 \\
% \hline

% Gauss                                   &4.78541 $\pm$ 0.0593234 \\
% Gauss + refinement                      &4.66389 $\pm$ 0.043216 \\
% \hline

% Laplace                                 &4.8479 $\pm$ 0.0385626 \\
% Laplace + refinement                    &4.80759 $\pm$ 0.0366873 \\
% \hline

% Softmax (CE + L2)                                 &4.7825 $\pm$ 0.0535609 \\
% Softmax (CE + L2) + refinement                    &4.65497 $\pm$ 0.0385003 \\
% \hline

% Softmax (CE + L2 + Var)                                 &4.806 $\pm$ 0.034842 \\
% Softmax (CE + L2 + Var) + refinement                    &4.68648 $\pm$ 0.034793 \\

Direct - Huber                        &4.80 $\pm$ 0.06 \\
Direct - Huber + Refinement           &4.74 $\pm$ 0.06 \\
\hline

Direct - L2                        &4.81 $\pm$ 0.02 \\
Direct - L2 + Refinement            &\textbf{4.65} $\pm$ 0.02 \\
\hline

Gaussian                                   &4.79 $\pm$ 0.06 \\
Gaussian + Refinement                      &4.66 $\pm$ 0.04 \\
\hline

Laplace                                 &4.85 $\pm$ 0.04 \\
Laplace + Refinement                    &4.81 $\pm$ 0.04 \\
\hline

Softmax - CE \& L2                                 &4.78 $\pm$ 0.05 \\
Softmax - CE \& L2 + Refinement                    &\textbf{4.65} $\pm$ 0.04 \\
\hline

Softmax - CE, L2 \& Var                                 &4.81 $\pm$ 0.03 \\
Softmax - CE, L2 \& Var + Refinement                    &4.69 $\pm$ 0.03 \\

\hline
\end{tabular}\vspace{-3mm}
\label{table:age_estimation_full}
\end{center}
\end{table}

\parsection{Laplace}
The Laplace model is defined using a DNN $f_\theta(x)$ according to,
\begin{equation}
\begin{gathered}
    p(y | x; \theta) = \frac{1}{2 \beta_{\theta}(x)} \exp{\big\{- \frac{|y - \mu_{\theta}(x)|}{\beta_{\theta}(x)} \big\}},\\
    f_{\theta}(x) = \rvect{\mu_{\theta}(x), \log\beta_{\theta}(x)}^{\Transp} \in \mathbb{R}^2.
\end{gathered}
\end{equation}
It is trained by minimizing the negative log-likelihood, corresponding to the loss,
\begin{equation}
    J(\theta) = \frac{1}{n} \sum_{i=1}^{n} \frac{|y_i - \mu_\theta(x_i)|}{\beta_{\theta}(x_i)} + \log\beta_{\theta}(x_i).
\end{equation}
The DNN architecture of $f_\theta(x)$ first extracts ResNet50 features $g_x \in \mathbb{R}^{2048}$ from the input image $x$. The feature vector $g_x$ is then processed by two heads of two fully-connected layers ($2048 \rightarrow 2048$, $2048 \rightarrow 1$) to output $\mu_{\theta}(x)$ and $\log\beta_{\theta}(x)$. The mean $\mu_{\theta}(x)$ is taken as the prediction $\hat{y}$. 

\parsection{Softmax}
The DNN architecture of \textit{Softmax} first extracts ResNet50 features $g_x \in \mathbb{R}^{2048}$ from the input image $x$. The feature vector $g_x$ is then processed by two fully-connected layers ($2048 \rightarrow 2048$, $2048 \rightarrow C$), outputting logits for $C = 101$ discretized classes $\{0, 1, \dots, 100\}$. It is trained by minimizing either the cross-entropy (CE) and $L^2$ losses, $J = J_{CE} + 0.1J_{L^2}$, or the CE, $L^2$ and variance~\cite{pan2018mean} losses, $J = J_{CE} + 0.1J_{L^2} + 0.05J_{Var}$. The prediction $\hat{y}$ is computed as the softmax expected value.

\subsection{Full Results}

Full experiment results, extending the results found in Table~4 (Section~4.3 in the paper), are provided in Table~\ref{table:age_estimation_full}.

\section{Head-Pose Estimation}
\label{appendix:head_pose_estimation}

In this appendix, further details on the head-pose estimation experiments (Section~4.4 in the paper) are provided.

\subsection{Network Architecture}
The DNN architecture $f_{\theta}(x,y)$ of our proposed model first extracts ResNet50 features $g_x \in \mathbb{R}^{2048}$ from the input image $x$. The pose $y \in \mathbb{R}^3$ is processed by four fully-connected layers (dimensions: $3 \rightarrow 16$, $16 \rightarrow 32$, $32 \rightarrow 64$, $64 \rightarrow 128$), generating $g_y \in \mathbb{R}^{128}$. The two feature vectors $g_x$, $g_y$ are then concatenated to form $g_{x,y} \in \mathbb{R}^{2048+128}$, which is processed by two fully-connected layers ($2048+128 \rightarrow 2048$, $2048 \rightarrow 1$), outputting $f_{\theta}(x, y) \in \mathbb{R}$.

\subsection{Training}
Our model is trained using $M = 1024$ samples from a proposal distribution $q(y | y_i)$ (Eq.~5 in the paper) with $L=2$ and variances $\sigma_1^2 = 1^2$, $\sigma_2^2 = 20^2$ for yaw, pitch and roll. An ablation study for the variances is found in Table~\ref{table:head_ablation}. The model is trained for $75$ epochs with a batch size of $32$, using the ADAM optimizer with weight decay of $0.001$. The images $x$ are of size $64 \times 64$. For data augmentation, we use random flipping along the vertical axis and random scaling in the range $[0.7, 1.4]$. After random flipping and scaling, a random image crop of size $64 \times 64$ is also selected. The ResNet50 is imported from \texttt{torchvision.models} in PyTorch with the pretrained option set to true, all other network parameters are randomly initialized using the default initializer in PyTorch.

\begin{table}[t]
\begin{center}
\caption{Impact of $\{\sigma_{l}\}_{l=1}^{2}$ in the proposal distribution $q(y|y_i)$~(Eq.~5 in the paper), for the head-pose estimation task on our validation split of the BIWI~\cite{fanelli2013random} dataset.}
\begin{tabular}{|l@{\hspace{1cm}}|l@{\hspace{1cm}}|}
\multicolumn{1}{c}{$\{\sigma_{l}\}_{l=1}^{2}$} &\multicolumn{1}{c}{Average MAE}\\ 
\hline
\{0.1, 20\} &6.96\\
\{1, 20\} &\textbf{5.08}\\
\{1, 30\} &5.24\\
\{2, 20\} &7.02\\
\{1, 10\} &7.56\\
\hline
\end{tabular}\vspace{-3mm}
\label{table:head_ablation}%
\end{center}
\end{table}

\subsection{Prediction}
For this experiment, we also use the prediction procedure detailed in Algorithm~\ref{algo:prediction_age}. Again following IoU-Net, we set $T = 5$, $\Omega_1 = 0.001$ and $\Omega_2 = -0.01$. Based on the validation set, we select $\lambda = 0.1$. We refine a single estimate $\hat{y}$, predicted by each baseline model.

\subsection{Baselines}
All baselines are trained for $75$ epochs with a batch size of $32$, using the ADAM optimizer with weight decay of $0.001$. Identical data augmentation and parameter initialization as for our proposed model is used.

\parsection{Direct}
The DNN architecture of \textit{Direct} first extracts ResNet50 features $g_x \in \mathbb{R}^{2048}$ from the input image $x$. The feature vector $g_x$ is then processed by two fully-connected layers ($2048 \rightarrow 2048$, $2048 \rightarrow 3$), outputting the prediction $\hat{y} \in \mathbb{R}^3$. It is trained by minimizing either the Huber or $L^2$ loss.

\parsection{Gaussian}
The Gaussian model is defined using a DNN $f_\theta(x)$ according to,
\begin{equation}
\begin{gathered}
   p(y | x; \theta) = \mathcal{N}\big(y; \mu_{\theta}(x), \Sigma_{\theta}(x)\big), \quad \Sigma_{\theta}(x) = \text{diag}\big(\sigma^2_{\theta}(x)\big),\\
   y = \rvect{y_1, y_2, y_3}^{\Transp} \in \mathbb{R}^3,\\
   \mu_{\theta}(x) = \rvect{\mu_{1,\theta}(x), \mu_{2, \theta}(x), \mu_{3, \theta}(x)}^{\Transp} \in \mathbb{R}^3,\\
   \sigma^2_{\theta}(x) = \rvect{\sigma^2_{1, \theta}(x), \sigma^2_{2, \theta}(x), \sigma^2_{3, \theta}(x)}^{\Transp} \in \mathbb{R}^3,\\
   f_{\theta}(x) = \rvect{\mu_{\theta}(x)^{\Transp}, \log\sigma^2_{\theta}(x)^{\Transp}}^{\Transp} \in \mathbb{R}^6.
\end{gathered}
\end{equation}
It is trained by minimizing the negative log-likelihood, corresponding to the loss,
\begin{equation}
    J(\theta) = \frac{1}{n} \sum_{i=1}^{n} \bigg( \sum_{k=1}^{3} \frac{(y_{k, i} - \mu_{k, \theta}(x_i))^2}{\sigma_{k, \theta}^{2}(x_i)} + \log \sigma_{k, \theta}^{2}(x_i) \bigg).
\end{equation}
The DNN architecture of $f_\theta(x)$ first extracts ResNet50 features $g_x \in \mathbb{R}^{2048}$ from the input image $x$. The feature vector $g_x$ is then processed by two heads of two fully-connected layers ($2048 \rightarrow 2048$, $2048 \rightarrow 3$) to output $\mu_{\theta}(x) \in \mathbb{R}^3$ and $\log\sigma^2_{\theta}(x) \in \mathbb{R}^3$. The mean $\mu_{\theta}(x)$ is taken as the prediction $\hat{y}$.

\begin{table*}[t]
\caption{Full results for the head-pose estimation experiments. Gradient-based refinement using our proposed method consistently improves the average MAE for yaw, pitch, roll (lower is better) for the predicted poses outputted by a number of baselines.}
\begin{center}
\begin{tabular}{|l||l|l|l|l|}
\multicolumn{1}{c}{Method} &\multicolumn{1}{c}{Yaw MAE} &\multicolumn{1}{c}{Pitch MAE} &\multicolumn{1}{c}{Roll MAE} &\multicolumn{1}{c}{Avg. MAE}\\ 
\hline

Yang et al. \cite{yang2018ssr}      &4.24    &4.35    &4.19    &4.26 \\
Gu et al. \cite{gu2017dynamic}         &3.91    &4.03    &3.03    &3.66 \\
Yang et al. \cite{yang2019fsa} &2.89    &4.29    &3.60    &3.60 \\
% SSR-Net-MD                          &4.24    &4.35    &4.19    &4.26 \\
% VGG16                               &3.91    &4.03    &3.03    &3.66 \\
% FSA-Caps-Fusion                     &2.89    &4.29    &3.60    &3.60 \\

\hline

Direct - Huber                  &2.78          $\pm$ 0.09   &3.73          $\pm$ 0.13    &2.90          $\pm$ 0.09    &3.14 $\pm$ 0.07 \\
Direct - Huber + Refine.      &2.75 $\pm$ 0.08   &3.70          $\pm$ 0.11    &2.87          $\pm$ 0.09    &3.11 $\pm$ 0.06 \\
\hline

Direct - L2                   &2.81 $\pm$ 0.08   &\textbf{3.60} $\pm$ 0.14    &2.85 $\pm$ 0.08    &3.09 $\pm$ 0.07 \\
Direct - L2 + Refine.      &2.78 $\pm$ 0.08   &3.62 $\pm$ 0.13    &2.81 $\pm$ 0.08    &3.07 $\pm$ 0.07 \\
\hline

Gaussian                               &2.89          $\pm$ 0.09   &3.64 $\pm$ 0.13    &2.83          $\pm$ 0.09    &3.12 $\pm$ 0.08 \\
Gaussian + Refine.                  &2.84          $\pm$ 0.08   &3.67          $\pm$ 0.12    &2.81 $\pm$ 0.08    &3.11 $\pm$ 0.07 \\
\hline

Laplace                             &2.93          $\pm$ 0.08   &3.80          $\pm$ 0.15    &2.90          $\pm$ 0.07    &3.21 $\pm$ 0.06 \\
Laplace + Refine.                &2.89          $\pm$ 0.07   &3.81          $\pm$ 0.13    &2.88          $\pm$ 0.06    &3.19 $\pm$ 0.06 \\
\hline

% Softmax                             &2.86          $\pm$ 0.12   &3.72          $\pm$ 0.25    &2.97          $\pm$ 0.11    &3.19 $\pm$ 0.10 \\
% Softmax + refinement                &2.81          $\pm$ 0.12   &3.72          $\pm$ 0.23    &2.94          $\pm$ 0.11    &3.16 $\pm$ 0.10 \\

Softmax - CE \& L2                             &2.73 $\pm$ 0.09   &3.63 $\pm$ 0.13    &2.77 $\pm$ 0.11    &3.04 $\pm$ 0.08 \\
Softmax - CE \& L2 + Refine.                &\textbf{2.67} $\pm$ 0.08   &3.61 $\pm$ 0.12    &\textbf{2.75} $\pm$ 0.10    &\textbf{3.01} $\pm$ 0.07 \\
\hline

Softmax - CE, L2 \& Var                             &2.83 $\pm$ 0.12   &3.79 $\pm$ 0.10    &2.84 $\pm$ 0.11    &3.15 $\pm$ 0.07 \\
Softmax - CE, L2 \& Var + Refine.                &2.76 $\pm$ 0.10   &3.74 $\pm$ 0.09    &2.83 $\pm$ 0.10    &3.11 $\pm$ 0.06 \\

\hline
\end{tabular}\vspace{-3.0mm}
\label{table:head_pose_estimation_full}
\end{center}
\end{table*}

\parsection{Laplace}
Following \cite{gast2018lightweight}, the Laplace model is defined using a DNN $f_\theta(x)$ according to,
\begin{equation}
\begin{gathered}
    p(y | x;\!\theta)\!=\!\prod_{k=1}^{3}\!\beta_{k, \theta}(x)^{\!-\!\frac{1}{2}} \exp{\!\big\{\!-\!\frac{1}{2}\!\bigg(\!\sum_{k=1}^{3}\!\frac{(y_k\!-\!\mu_{k, \theta}(x))^2}{\beta_{k, \theta}(x)}\!\bigg)^{\frac{1}{2}}\!\big\}}\!,\\
    y = \rvect{y_1, y_2, y_3}^{\Transp} \in \mathbb{R}^3,\\
    \mu_{\theta}(x) = \rvect{\mu_{1,\theta}(x), \mu_{2, \theta}(x), \mu_{3, \theta}(x)}^{\Transp} \in \mathbb{R}^3,\\
    \beta_{\theta}(x) = \rvect{\beta_{1, \theta}(x), \beta_{2, \theta}(x), \beta_{3, \theta}(x)}^{\Transp} \in \mathbb{R}^3,\\
    f_{\theta}(x) = \rvect{\mu_{\theta}(x)^{\Transp}, \log\beta_{\theta}(x)^{\Transp}}^{\Transp} \in \mathbb{R}^6.
\end{gathered}
\end{equation}
It is trained by minimizing the negative log-likelihood, corresponding to the loss,
\begin{equation}
    J(\theta)\!=\!\frac{1}{n}\!\sum_{i=1}^{n}\!\bigg\{\!\bigg(\!\sum_{k=1}^{3}\!\frac{(y_{k, i}\!-\!\mu_{k, \theta}(x_i))^2}{\beta_{k, \theta}(x_i)}\!\bigg)^{\frac{1}{2}}\!+\!\sum_{k=1}^{3} \log \beta_{k, \theta}(x_i)\!\bigg\}\!.
\end{equation}
The DNN architecture of $f_\theta(x)$ first extracts ResNet50 features $g_x \in \mathbb{R}^{2048}$ from the input image $x$. The feature vector $g_x$ is then processed by two heads of two fully-connected layers ($2048 \rightarrow 2048$, $2048 \rightarrow 3$) to output $\mu_{\theta}(x) \in \mathbb{R}^3$ and $\log\beta_{\theta}(x) \in \mathbb{R}^3$. The mean $\mu_{\theta}(x)$ is taken as the prediction $\hat{y}$.

\parsection{Softmax}
The DNN architecture of \textit{Softmax} first extracts ResNet50 features $g_x \in \mathbb{R}^{2048}$ from the input image $x$. The feature vector $g_x$ is then processed by three heads of two fully-connected layers ($2048 \rightarrow 2048$, $2048 \rightarrow C$), outputting logits for $C = 151$ discretized classes $\{-75, -74, \dots, 75\}$ for the yaw, pitch and roll angles (in degrees). It is trained by minimizing either the cross-entropy (CE) and $L^2$ losses, $J = J_{CE} + 0.1J_{L^2}$, or the CE, $L^2$ and variance~\cite{pan2018mean} losses, $J = J_{CE} + 0.1J_{L^2} + 0.05J_{Var}$. The prediction $\hat{y}$ is obtained by computing the softmax expected value for yaw, pitch and roll.

\subsection{Full Results}

Full experiment results, extending the results found in Table~5 (Section~4.4 in the paper), are provided in Table~\ref{table:head_pose_estimation_full}.
\end{appendices}

\end{document}